%% file: neurips_2022.tex
\UseRawInputEncoding
\pdfoutput=1
\documentclass{article}

    \usepackage[final]{neurips_2022}

\usepackage[utf8]{inputenc} 
\usepackage[T1]{fontenc}    
\usepackage{hyperref}       
\usepackage{url}            
\usepackage{booktabs}       
\usepackage{amsfonts}       
\usepackage{nicefrac}       
\usepackage{microtype}      
\usepackage{xspace}
\usepackage{graphicx}
\usepackage{amsmath}
\usepackage{wrapfig}
\usepackage{makecell}
\usepackage{subfigure}
\newtheorem{theorem}{Theorem}
\newtheorem{lemma}{Lemma}
\newtheorem{proposition}{Proposition}

\usepackage{tabularx}
\usepackage{soul}
\usepackage{graphicx}

\usepackage{eucal}
\usepackage{tabularx}

\usepackage{color,soul}
\usepackage{caption}
\usepackage{float} 
\usepackage[normalem]{ulem}
\usepackage[dvipsnames]{xcolor}
\usepackage[ruled,vlined,linesnumbered,lined,boxed,commentsnumbered]{algorithm2e}

\newcommand{\ourmethod}{\texttt{GGD}}
\newcommand{\change}{\textcolor{black}}

\title{Rethinking and Scaling Up Graph Contrastive Learning: An Extremely Efficient Approach with Group Discrimination}

\author{
  Yizhen Zheng$^1$,\ Shirui Pan$^2$\thanks{Corresponding Author.}, \  Vincent CS Lee$^1$, \ Yu Zheng$^3$,\ Phillip S. Yu$^4$, \\
  $^1$Monash University, $^2$Griffith University, $^3$La Trobe University, $^4$ University of Illinons at Chicago\\
  \texttt{yizhen.zheng1@monash.edu, s.pan@griffth.edu.au, vincent.cs.lee@monash.edu} \\ \texttt{yu.zheng@latrobe.edu.au, psyu@uic.edu} 
}

\begin{document}

\maketitle

\begin{abstract}
\label{sec:abstract}
\input{1_Abstract}
\end{abstract}

\section{Introduction}
\label{sec:intro}
\input{2_Introduction}

\section{Rethinking Representative GCL Methods}
\label{sec:rethink}
\input{3_Rethinking_GCL}

\section{Methodology}
\label{sec: method}
\input{4_Method}


\section{Related Work}
\label{sec: related_work}
\input{6_Related_Work}

\section{Experiments}
\label{sec:experiment}
\input{7_Experiment}

\section{Explore Group Discrimination}
\label{sec:explore gd}
\input{10_Explore_GD}

\section{Future Work}
\label{sec: future work}
\input{8_Future_work}


{\small
\bibliographystyle{unsrt}
\bibliography{neurips_2022}}

\section*{Checklist}

\input{9_checklist}

\appendix

\input{Appendix_A}

\end{document}

%% file: 1_Abstract.tex
Graph contrastive learning (GCL) alleviates the heavy reliance on label information for graph representation learning (GRL) via self-supervised learning schemes. The core idea is to learn by maximising  mutual information for similar instances, which requires similarity computation between two node instances. However, GCL is inefficient in both time and memory consumption.
In addition, GCL normally requires a large number of training epochs to be well-trained on large-scale datasets. Inspired by an observation of a technical defect (i.e., inappropriate usage of Sigmoid function) commonly used in two representative GCL works, DGI and MVGRL, we revisit GCL and introduce a new learning paradigm for self-supervised graph representation learning, namely, Group Discrimination (GD), and propose a novel GD-based method called \underline{G}raph \underline{G}roup \underline{D}iscrimination (\ourmethod). Instead of similarity computation, \ourmethod\  directly discriminates two groups of \change{node samples with a very simple binary cross-entropy loss.}
In addition, \ourmethod\ requires much \change{fewer} training epochs to obtain competitive performance compared with GCL methods on large-scale datasets. These two advantages endow \ourmethod\ with very efficient property. Extensive experiments show that \ourmethod\  outperforms state-of-the-art self-supervised methods on \textit{eight} datasets. In particular, \ourmethod\ can be trained in 0.18 seconds (6.44 seconds including data preprocessing) on ogbn-arxiv, which is \textbf{orders of magnitude (\textbf{10,000+}) faster than GCL baselines}  while consuming much less memory. Trained with 9 hours on ogbn-papers100M with billion edges, \ourmethod\ outperforms its GCL counterparts in both accuracy and efficiency. 

%% file: 2_Introduction.tex
Graph Neural Networks (GNNs) have been widely-adopted in learning representations for graph-structured data.
By utilising message-passing over the topology of a graph,
GNNs can learn effective low-dimensional node embeddings, which can be used for a variety of downstream tasks such as node classification~\cite{jin2021multi}. GNNs have been further applied in diverse domains, e.g., federated learning~\cite{tan2022federated, tan2022fedproto}, trustworthy systems~\cite{zhang2022trustworthy,zhang2021projective}, dynamic graphs~\cite{jin2021neural, jin2022multivariate} and anomaly detection~\cite{liu2021anomaly, zheng2021heterogeneous}.

\begin{wrapfigure}{r}{0.58\textwidth}
\vspace{-2mm}
    \centering
    \includegraphics[scale = 1.1]{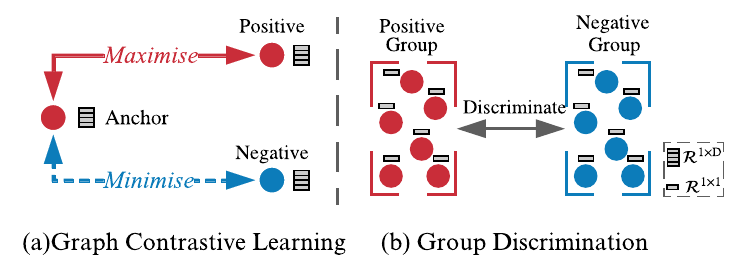}
    \caption{The left subfigure shows the GCL learning scheme. Red line indicates MI maximisation between two  nodes, each of which $ \in \mathbb{R}^{1 \times D}$, while blue line indicates the opposite operation. The right subfigure presents Group Discrimination. It discriminates \change{positive and negative node samples,  each of which $ \in \mathbb{R}^{1 \times 1}$.}}
    \label{fig:node_comparison}
    \vspace{-6mm}
\end{wrapfigure}

However, many GNNs adopt a supervised learning manner to train models with label information, which is expensive and labour-intensive to collect in real-world. To address this issue, a few studies (e.g., DGI~\cite{velivckovic2018deep}, MVGRL~\cite{hassani2020contrastive}, GMI~\cite{peng2020graph}, and GRACE~\cite{zhu2020deep}) borrow the idea of contrastive learning from computer vision (CV), and introduce graph contrastive learning (GCL) methods for self-supervised GRL. The core idea of these methods is to maximise the mutual information (MI) between an anchor node and its positive counterparts, sharing similar semantic information while doing the opposite for negative counterparts as shown in Figure \ref{fig:node_comparison}(a). Nonetheless, such a scheme relies on similarity calculation in contrastive loss computation. 
Additionally, GCL normally requires a large number of training \change{epochs} to be well-trained on large-scale datasets. Thus, when the size of the dataset is large, these methods require a significant amount of time and resources to be well-trained.

\begin{wraptable}{r}{0.5\textwidth}
\vspace{-0.4cm}
\footnotesize
  \caption{Training time in \underline{seconds} comparison between \ourmethod\ and GBT~\cite{bielak2021graph} (i.e., the most efficient GCL baseline as shown in section \ref{sec:evalsmall}) on ogbn-arxiv. Number in brackets means the hidden size. `Pre', `Tr' and `Epo' indicate preprocessing time, training time per epoch,
   and the number of epochs for training GNNs. `Total(E)' and `Total(T)' are total end-to-end training time (i.e., including preprocessing), which equals to ($\rm Pre + Epo \times Tr$) and total training time, which is ($\rm Epo \times Tr$). `Imp(E)' and `Imp(T)' indicate how many times \ourmethod\ improve on `Total(E)' and `Total(T)'. `Acc' is averaged accuracy result on test set over five runs. 
  }
  \resizebox{0.5\columnwidth}{!}{
  \begin{tabularx}{1.5\linewidth}{p{1.5cm}<{\centering}|p{0.3cm}<{\centering} p{0.3cm}<{\centering} p{0.4cm}<{\centering}p{0.7cm}<{\centering}p{1.2cm}<{\centering} p{0.6cm}<{\centering} p{1.22cm}<{\centering}|p{0.3cm}<{\centering}}
    \toprule
    \textbf{Method} & \textbf{Pre} &\textbf{Tr} & \textbf{Epo} & \textbf{Total(E)}  & \textbf{Imp(E)} &  \textbf{Total(T)} & \textbf{Imp(T)} & \textbf{Acc} \\
    \midrule
    GBT(256) & 5.52  & 6.47 & 300 & 1,946.52 & - & 1,941.00 & - & 70.1  \\
    \midrule
    \ourmethod (256)  & 6.26 & 0.18 & 1 &  6.44 &  \textbf{302.25 $\times$} & 0.18 &\textbf{\color{red}{10,783.33$\times$}} & 70.3\\
    \ourmethod (1,500)  & 6.26 & 0.95 & 1 & 7.21  & \textbf{269.96$\times$} & 0.95 & \textbf{2,043.16$\times$} & 71.6\\
  \bottomrule
\end{tabularx}}
\label{tab:summary ogbn-arxiv}
\vspace{-2mm}
\end{wraptable}

\change{Though a few GCL works attempt to improve graph contrastive learning with specially designed schemes, e.g., BGRL~\cite{thakoor2021bootstrapped} and GBT~\cite{bielak2021graph}, they are still inefficient and require high time consumption for model training. Inspired by BYOL~\cite{grill2020bootstrap}}, BGRL~\cite{thakoor2021bootstrapped} adopts a bootstrapping scheme and remove negative node pairs. It only contrasts a node from the online network (i.e., updated with gradient) to its corresponding embedding from the target network (i.e., updated momentumly with stop gradient). Based on Barlow-Twins~\cite{zbontar2021barlow}, GBT~\cite{bielak2021graph} \change{borrows the idea of redundancy-reduction principle and utilises a cross-correlation-based loss to build contrastiveness between embedding dimensions.}


To boost training efficiency of self-supervised GRL, inspired by an observation of a technical defect (i.e., inappropriate application of Sigmoid function) in two representative GCL studies, we introduce a novel learning paradigm, namely, \underline{G}roup \underline{D}iscrimination (GD). 
Instead of similarity computation, GD directly discriminates a group of positive nodes from a group of negative nodes, as shown in Figure \ref{fig:node_comparison}(b). Specifically, GD defines \change{node samples} generated with original 
graph as the positive group, while \change{node samples} obtained with corrupted topology are regarded as the negative group. Then, GD trains the model by classifying these \change{node samples} into the correct group with a very simple binary cross-entropy loss.
By doing so, the model can extract valuable self-supervised signals from learning the edge distribution of a graph.
Compared with GCL, GD enjoys numerous merits including extremely fast training, fast convergence (e.g., 1 epoch to be well-trained on large-scale datasets), and high scalability while achieving SOTA performance with existing GCL approaches.  

Using GD as backbone, we design a new self-supervised GRL model with the Siamese structure called \underline{G}raph \underline{G}roup \underline{D}iscrimination (GGD). 
Firstly, we can optionally augment a given graph with augmentation techniques, e.g., feature and edge dropout. Then, the augmented graph \change{is} fed into a GNN encoder and a projector to obtain embeddings for the positive group. After that, the augmented feature is corrupted with node shuffling (i.e., disarranging the order of nodes in the feature matrix) to disrupt the topology of a graph and input to the same network for obtaining embeddings of the opposing group. Finally, the model is trained by discriminating \change{these two groups of node samples}. 
The contributions of this paper are \change{three-fold}: 1) We re-examine existing GCL approaches (e.g., DGI \cite{velivckovic2018deep} and MVGRL \cite{hassani2020contrastive}), and we introduce a novel and efficient self-supervised GRL paradigm, namely, Group Discrimination (GD). 
2) Based on GD, we propose a new self-supervised GRL model, GGD, which is fast in training and convergence, and possess
high scalability. 3) We conduct extensive experiments on eight datasets, including an extremely large dataset, ogbn-papers100M with billion edges. The experiment results show that our proposed method \textbf{reaches state-of-the-art performance while consuming much less time and memory than baselines, e.g., 10783 $\times$ faster} than the most efficient GCL baseline with its best selected \change{epochs} number \cite{bielak2021graph}, as shown in Table \ref{tab:summary ogbn-arxiv}.

    
    

%% file: 3_Rethinking_GCL.tex
\begin{wrapfigure}{r}{0.5\textwidth}
\vspace{-4.5mm}
    \centering
    \includegraphics[scale = 0.7]{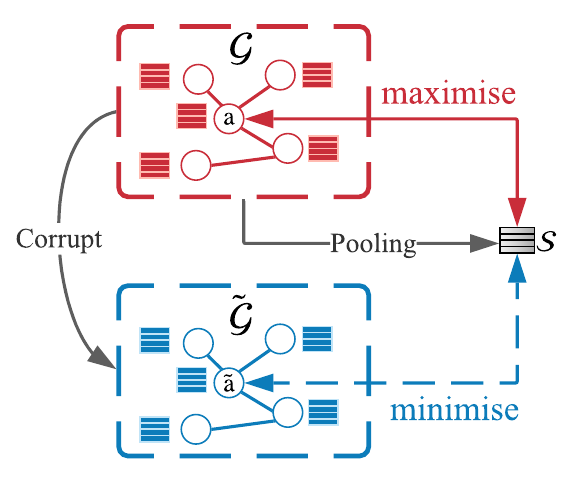}
    \caption{The architecture of DGI. Cubes indicate node embeddings. Red and blue lines represent MI maximisation and minimisation, respectively. $\mathcal{G}$ and $\tilde{\mathcal{G}}$ denote the original graph and the corrupted graph. $\textbf{s}$ is the summary vector.
    }
    \label{fig:DGI}
    \vspace{-10mm}
\end{wrapfigure}
In this section, we analyse a technical defect observed in two representative GCL methods, DGI~\cite{velivckovic2018deep} and MVGRL~\cite{hassani2020contrastive}. Based on the technical defect, we show that  mutual information maximisation behind these two approaches is not the contributed factor to contrastive learning, but a new paradigm, group discrimination. Finally, from the analysis, we provide the definition of this new concept.


\subsection{Rethinking GCL Methods}
\label{rethink_gcl}
DGI~\cite{velivckovic2018deep} is the first work introducing contrastive learning into GRL. However, due to a technical defect observed in their official open-source code, we found it is essentially not working as the authors thought (i.e., learning via MI interaction). 


\vspace{1mm}\noindent\textbf{Constant Summary Vector.} As shown in Figure \ref{fig:DGI}, the original idea of DGI is to maximise the MI (i.e., the red line) between a node $a$ and the summary vector $\textbf{s}$, which is obtained by averaging all node embeddings in a graph $\mathcal{G}$. Also, to regularise the model training, DGI corrupts $\mathcal{G}$ by shuffling the node order of the input feature matrix to get $\tilde{\mathcal{G}}$. Then, generated embeddings of $\tilde{\mathcal{G}}$ \change{serve} as negative samples, which are pulled apart from the summary vector $\textbf{s}$ via MI minimisation.


\begin{wraptable}{r}{0.6\textwidth}
\vspace{-3mm}
\footnotesize
\centering
  \caption{Summary vector statistics on three datasets with different activation functions including ReLU, LeakyReLU (i.e., LReLU shown below), PReLU, and Sigmoid.}
  \label{tab:summary vector}
  \resizebox{0.6\columnwidth}{!}{
  \change{
  \begin{tabularx}{1.5\linewidth}{l|p{2.0cm}p{2.0cm}<{\centering}p{2.0cm}<{\centering} p{2.0cm}<{\centering}p{2.0cm}<{\centering}p{2.0cm}<{\centering}}
    \toprule
    \textbf{Activation} & \textbf{Statistics} &\textbf{Cora} & \textbf{CiteSeer} & \textbf{PubMed}\\
    \midrule
     & Mean & 0.50 & 0.50 & 0.50\\
    ReLU/LReLU/PReLU &Std & 1.3e-03 & 1.0e-04 & 4.0e-04\\
     & Range & 1.4e-03 & 8.0e-04 & 1.5e-03 \\
  \midrule
    & Mean & 0.62 & 0.62 & 0.62 \\
    Sigmoid & Std & 5.4e-05 & 2.9e-05 & 6.6e-05\\
    & Range & 3.6e-03 & 3.0e-03 & 3.2e-03 \\
  \bottomrule
\end{tabularx}}}
\label{tab:summ stat}
\end{wraptable}

Nonetheless, in the implementation of DGI, a Sigmoid function is inappropriately applied on the summary vector generated from a GNN whose weight is initialised with Xavier initialisation.
As a result, elements in the summary vector are very close to the same value. We have validated this finding on three datasets, Cora, CiteSeer and PubMed. The experiment result is shown in Table \ref{tab:summary vector}, which shows that summary vectors in all datasets are approximately a \textbf{constant vector} $\epsilon \textbf{\textit{I}}$, where $\epsilon$ is a scalar and $ \textbf{\textit{I}}$ is an all-ones vector (\change{i.e., $\epsilon$=0.50 with ReLU/LReLU/PReLU and $\epsilon$=0.62 with Sigmoid as non-linear activation in these datasets}).

\begin{wraptable}{r}{0.43\textwidth}
\vspace{-5mm}
\footnotesize
  \caption{The experiment result on three datasets with changing value from 0 to 1.0 for the summary vector.}
  \label{tab:fix vector}
  \resizebox{0.45\columnwidth}{!}{
  \begin{tabularx}{1.55\linewidth}{lp{0.9cm}<{\centering} p{0.9cm}<{\centering} p{0.9cm}<{\centering} p{0.9cm}<{\centering} p{0.9cm}<{\centering} p{0.9cm}<{\centering}}
    \toprule
    \textbf{Dataset} &\textbf{0} & \textbf{0.2} & \textbf{0.4}  & \textbf{0.6}  & \textbf{0.8}  & \textbf{1.0}\\
    \midrule
    Cora & 70.3$\pm{0.7}$ & 82.4$\pm{0.2}$ &  82.3$\pm{0.3}$ & 82.5$\pm{0.4}$  & 82.3$\pm{0.3}$ & 82.5$\pm{0.1}$ \\
    CiteSeer & 61.8$\pm{0.8}$  & 71.7$\pm{0.6}$ & 71.9$\pm{0.7}$ & 71.6$\pm{0.9}$ & 71.7$\pm{1.0}$ & 71.6$\pm{0.8}$\\
    PubMed & 68.3$\pm{1.5}$ & 77.8$\pm{0.5}$ & 77.9$\pm{0.8}$ & 77.7$\pm{0.9}$ & 77.4$\pm{1.1}$ & 77.2$\pm{0.9}$ \\
  \bottomrule
\end{tabularx}}
\vspace{-8mm}
\end{wraptable}
To theoretically explain this phenomenon, we present the proposition below:
\begin{proposition}
Given $\normalfont{\mathcal{G} = \{\textbf{X} \in \mathbb{R}^{N\times D}, \textbf{A} \in \mathbb{R}^{N\times N}\}}$, and a GCN encoder $g(\cdot)$ initialised with Xavier initialisation, we can obtain its embedding $\normalfont{\textbf{H} = \sigma(g(\mathcal{G}))}$, where $\sigma(\cdot)$ is a non-linear activation function. By applying the sigmoid function $\sigma_{sig}(\cdot)$ to the summary vector $\normalfont{\textbf{s}}$ (i.e., the average row vector of $\normalfont{\textbf{H}}$), values in $\sigma_{sig}(\normalfont{\textbf{s}})$ approximately become 0.5 with ReLU/LReLU/PReLU or 0.62 with Sigmoid as non-linear activation of $g(\cdot)$ at the initialisation stage.
\label{th:th1}
\end{proposition}
Based on this proposition, we can see these summary vectors can lose variance and become a constant vector at the initialisation stage. Based on Table \ref{tab:summ stat}, we can see the constant in the summary vector remain unchanged, and the information loss still occurs even if the GNN encoder is trained. Thus, we conjecture the training process won't affect the constant value much in the summary vector of DGI.
The proof for \change{the proposition} is presented in Appendix \ref{proof 1}.



To evaluate the effect of $\epsilon$ to constant summary vector, we vary the scalar $\epsilon$ (from  0 to 1 increment by 0.2) to change the constant summary vector 
and report the model performance (i.e., averaged accuracy on five runs) in Table \ref{tab:fix vector}.

From this table, we can see, except for 0, the model performance is trivially affected by $\epsilon$ for constant summary vector. When the summary vector is set to 0, the model performance plummets because node embeddings become all 0 when multiplying with such vector and the model converges to the trivial solution. As the summary vector only has a trivial effect on model training, the hypothesis of DGI~\cite{velivckovic2018deep} on learning via contrastiveness between anchor nodes and the summary instance does not hold, which raises a question to be investigated: \textit{What truly leads to the success of DGI?}

\noindent\noindent\textbf{Simplifying DGI.} To answer the question, we predigest the objective function proposed in DGI (i.e., maximising the MI between $\textbf{h}_i$ and the summary vector $\textbf{s}$) by using an all-ones vector as the summary vector $\textbf{s}$ (i.e., setting $\textbf{s}=\epsilon  \textbf{\textit{I}}=\textbf{\textit{I}}$) and simplifying the discriminator $\mathcal{D}(\cdot)$ (i.e., removing the learnable weight matrix). Then, we rewrite the objective function to the following form:
\begin{equation}
    \begin{aligned}
     \mathcal{L}_{DGI} &=  \frac{1}{2N}({\sum_{i = 1}^{N}}\log\mathcal{D}(\textbf{h}_i, \textbf{s}) +\log(1 - \mathcal{D}(\tilde{\textbf{h}}_i, \textbf{s}))), \\
    &  = \frac{1}{2N}(\sum_{i = 1}^{N}\log(\textbf{h}_i \cdot \textbf{s}) + \log(1 - \tilde{\textbf{h}}_i\cdot \textbf{s}))), \\
    &  = \frac{1}{2N}({\sum_{i = 1}^{N}}\log(sum(\textbf{h}_i)) + \log(1 - sum(\tilde{\textbf{h}}_i))), 
    \end{aligned}
    \label{eq:DGI loss}
\end{equation}
where $\cdot$ is the vector multiplication operation, $N$ is the number of nodes in a graph, $ \textbf{h}_i \in \mathbb{R}^{1 \times D}$ and $ \tilde{\textbf{h}}_i \in \mathbb{R}^{1 \times D}$ are the original and corrupted embedding for node $i$, 
$ sum(\cdot)$ is the summation function, and $\mathcal{D}(\cdot)$ is a discriminator for bilinear transformation, which can be formulated as follows: 
\begin{equation}
     \mathcal{D}(\textbf{h}_i, \textbf{s}) = \sigma_{sig}(\textbf{h}_i\cdot \textbf{W}\cdot \textbf{s}),
    \label{eq:discriminator}
\end{equation}
\begin{wraptable}{r}{0.45\textwidth}
\vspace{-4mm}
\footnotesize
  \caption{Comparison of the original DGI and $ \text{DGI}_{BCE}$ in terms of accuracy (averaged on five runs), memory efficiency (in MB) and training time (in seconds). Number after | shows how many times have $ \text{DGI}_{BCE}$ improved on top of DGI.}
  \label{tab:bce_compare}
  \resizebox{0.45\columnwidth}{!}{
  \begin{tabularx}{1.55\linewidth}{ll p{1.6cm}<{\centering} p{1.6cm}<{\centering}p{1.6cm}<{\centering}}
    \toprule
    \textbf{Experiment} & \textbf{Method} &\textbf{Cora} & \textbf{CiteSeer} & \textbf{PubMed}\\
    \midrule
    Accuracy  & DGI & 81.7$\pm{0.6}$ & 71.5$\pm{0.7}$ & 77.3$\pm{0.6}$ \\
     & $ \text{DGI}_{BCE}$ & 82.5$\pm{0.3}$ & 71.7$\pm{0.6}$ & 77.7$\pm{0.5}$\\
    \midrule
    Memory & DGI & 4189MB & 8199MB & 11471MB \\
     & $ \text{DGI}_{BCE}$ & 1475MB|64.8\% & 1587MB|80.6\% & 1629MB|85.8\% \\
    \midrule
    Time & DGI & 0.085s & 0.134s & 0.158s \\
     & $ \text{DGI}_{BCE}$ & 0.010s|8.5$\times$ & 0.021s|6.4$\times$ & 0.015s|10.5$\times$\\
  \bottomrule
\end{tabularx}}
\vspace{-2mm}
\end{wraptable}
where $ \textbf{W}$ is a learnable weight matrix and $\sigma_{sig}(\cdot)$ is the sigmoid function. Specifically, as shown in Equation \ref{eq:discriminator}, by removing the weight matrix $ \textbf{W}$, $ \textbf{h}_i$ is directly multiplied with $ \textbf{s}$. As $ \textbf{s}$ is a vector containing only one, the multiplication of $ \textbf{h}_i$ and $ \textbf{s}$ is equivalent to summing $ \textbf{h}_i$ itself directly. From this form, we can see that the multiplication of $ \textbf{h}_i$ and the summary vector only serves as an aggregation function (i.e., summation aggregation) to \change{aggregate} $ \textbf{h}_i$. To explore the effect of other aggregation functions, we replace the summation function in Equation \ref{eq:DGI loss} with other aggregation methods such as mean-, minimum-, and maximum- pooling, and present the experiment result in Appendix \ref{ap:aggregation}. 


Based on Equation \ref{eq:DGI loss}, we can rewrite it to a very simple binary cross entropy loss if we also include corrupted nodes as data samples and setting \change{$\hat{y}_i = agg(\textbf{h}_i)$, where $agg(\cdot)$ stands for aggregation}:
\begin{equation}
     \mathcal{L}_{BCE} =- \frac{1}{2N}(\sum_{i = 1}^{2N}y_i\log \hat{y}_i + (1 - y_i)\log(1 - \hat{y}_i)),
    \label{eq:binary DGI}
\end{equation}
where $ y_i \in \mathbb{R}^{1 \times 1}$ means the indicator for node $i$ (i.e., if node $i$ is corrupted, $ y_i$ is 0, otherwise it is 1), and $ \hat{y}_i \in \mathbb{R}^{1 \times 1}$ represents the \change{prediction for a node sample $i$}. As we include corrupted nodes as data samples, the size of nodes to be processed is doubled to $ 2N$ (i.e., the number of corrupted nodes is equal to the number of original nodes). From the equation above, we can easily observe that what DGI truly does is discriminate between a group of \change{nodes generated with correct topology and nodes generated with corrupted topology}, as shown in Figure \ref{fig:node_comparison}. We name this self-supervised learning paradigm "\textbf{Group Discrimination}". To validate the effectiveness of this paradigm, we replace the original DGI loss with Equation \ref{eq:binary DGI}, namely, $ \text{DGI}_{BCE}$ and compare it with DGI on three datasets in terms of training time, memory efficiency and model performance as shown in Table \ref{tab:bce_compare}. Here, $ \text{DGI}_{BCE}$ adopts the same parameter setting as DGI. From this table, we can observe $ \text{DGI}_{BCE}$ dramatically improves DGI in both memory and time efficiency while it slightly enhances the model performance of DGI. This may be contributed to the removal of multiplication operations between node pairs, which eases the burden of computation and memory consumption. 

Similar to DGI, the same technical defect is observed in MVGRL~\cite{hassani2020contrastive}, which makes it become a GD-based method. Extended on DGI, MVGRL~\cite{hassani2020contrastive} incorporates diffusion augmentation to inject additional global information \change{into} model training, which enhances the model performance. The detailed analysis for MVGRL is presented in Appendix \ref{ap: rethink mvgrl}.




\subsection{Definition of Group Discrimination}
As mentioned above, Group Discrimination is a self-supervised GRL paradigm, which learns by discriminating different groups of \change{node samples}. Specifically, the paradigm assigns different indicators to different groups of \change{node samples}. For example, for binary group discrimination, one group is considered as the positive group with class 1 as its indicator, whereas the other group is the negative group, having its indicator assigned as 0. Given a graph $\mathcal{G}$, the positive group usually includes node \change{samples} generated with the original graph $\mathcal{G}$ or its augmented views (i.e.,  similar graph instances of $\mathcal{G}$ created by augmentation). In contrast, the opposing group contains negative samples obtained by corrupting $\mathcal{G}$, e.g., changing its topology structure. 

\change{Based on our theoretical analysis, group discrimination is learning to avoid making `mistakes' (i.e., bias the encoder towards avoiding mistaken samples), thus improving the quality of generated embeddings. The analysis and an intuitive explanation are presented in Section \ref{ap:theory} and \ref{connection}}.

%% file: 4_Method.tex
We first define unsupervised node representation learning and then present the architecture of \ourmethod, which extends $\textbf{DGI}_{BCE}$ with additional augmentation, \change{the projector} and embedding reinforcement to reach better model performance. Given a graph $\mathcal{G}$ with attributes $ \textbf{X} \in \mathbb{R}^{N \times D}$, where $ N$ is the number of nodes in $\mathcal{G}$, and $ D$ is the number of dimensions of $\textbf{X}$, our aim is to train a GNN encoder without the reliance on labelling information. With the trained encoder, taking $\mathcal{G}$ and $\textbf{X}$ as input, it can output learned representations $ \textbf{H} \in \mathbb{R}^{N \times {D'}}$, where $ D'$ is the predefined hidden dimension. $ \textbf{H}$ can then be used in many downstream tasks such as node classification.

\begin{figure*}[ht]
    \centering
    \includegraphics[scale = 0.85]{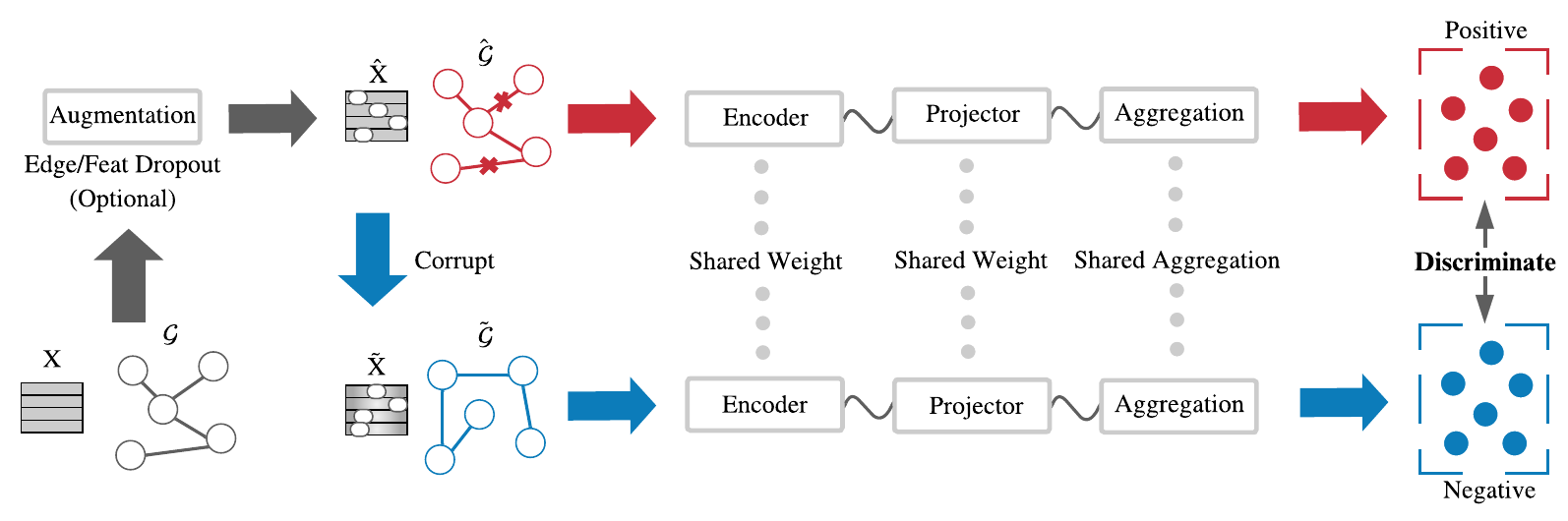}
    \caption{The architecture of \ourmethod. Given a graph $\mathcal{G}$ \change{with} a feature matrix $ \textbf{X}$, we can optionally apply augmentation on them to generate  $\hat{\mathcal{G}}$ and $ \hat{\textbf{X}}$. Then, we corrupt $ \hat{\textbf{X}}$ and $\hat{\mathcal{G}}$ to obtain $ \tilde{\textbf{X}}$ and $\tilde{\mathcal{G}}$. Taking $ \hat{\textbf{X}}$ and $\hat{\mathcal{G}}$ as input to the encoder and the projector\change{, i.e., a multilayer perceptron}, positive node samples can be obtained. Similarly, $\tilde{\textbf{X}}$ and $\tilde{\mathcal{G}}$ are fed to the same encoder and projector to generate negative samples. \change{The generated embeddings are aggregated to get predictions for the group discrimination task. This process will be iteratively conducted until reaching the predefined training epochs.}}
    \label{fig:ggd}
    \vspace{-2mm}
\end{figure*}

\subsection{Graph Group Discrimination}
Based on the proposed self-supervised GRL paradigm, group discrimination, we have designed a novel method, namely \ourmethod, to learn node representations using a siamese network structure and a \change{BCE} loss. The architecture of  \ourmethod\ is presented in Figure \ref{fig:ggd}. The framework mainly consists of four components: augmentation, corruption, a siamese GNN network, and group discrimination.

\noindent \textbf{Augmentation.} With a given graph $\mathcal{G}$ and feature matrix $ \textbf{X}$, optionally, we can augment it with augmentation techniques such as edge and feature dropout to create $\hat{\mathcal{G}}$ and $ \hat{\textbf{X}}$. In practice, we follow the augmentation proposed in GraphCL~\cite{you2020graph}. Specifically, edge dropout removes a predefined fraction of edges, while we use node dropout to mask a predefined proportion of feature dimension, i.e., assigning 0 to replace values in randomly selected dimensions. This step is optional in implementation.

Notably, the motivation of using augmentation in our framework is distinct from contrastive learning methods. In our study, augmentation is used to increase the difficulty of the self-supervised training tasks. With augmentation, $\hat{\mathcal{G}}$ and $ \hat{\textbf{X}}$ \change{change} in every training iteration, which forces the model to lessen the dependence on the fixed pattern (i.e., unchanged edge and feature distribution) in a monotonous graph. However, in contrastive learning, augmentation creates augmented views sharing similar semantic information for building contrastiveness. 

\noindent \textbf{Corruption.} $\hat{\mathcal{G}}$ and $ \hat{\textbf{X}}$ are then corrupted to build $\tilde{\mathcal{G}}$ and $ \tilde{\textbf{X}}$ for the generation of node embeddings in the negative group. We adopt the same corruption technique used in DGI~\cite{velivckovic2018deep} and MVGRL~\cite{hassani2020contrastive} (as shown in Figure \ref{fig:corruption}).
The corruption technique devastates the topology structure of  $\hat{\mathcal{G}}$ by randomly changing the order of nodes in $ \hat{\textbf{X}}$. The corrupted $ \tilde{\textbf{X}}$ and $\tilde{\mathcal{G}}$ can be used for producing node representations with incorrect network connections.



\noindent \textbf{The Siamese GNN.} We have designed a siamese GNN network to output node representations given a graph and its attribute. The siamese GNN network is made up of two components, which are a GNN encoder and a projector. The backbone GNN encoder is replaceable with a variety of choices of GNNs, e.g., GCN~\cite{kipf2016semi} and GAT~\cite{velivckovic2017graph}. In our work, we adopt GCN as the backbone. The projector is a \change{multi-layer perceptron network, whose number of layers can be adjusted.} When generating node embeddings of the positive group, the Siamese network takes $\hat{\mathcal{G}}$ and $ \hat{\textbf{X}}$ as input. Using the same encoder and projector, the Siamese network output the negative group with $\tilde{\mathcal{G}}$ and $ \tilde{\textbf{X}}$. These two groups of node embeddings are considered as a collection of data samples with a size of $ 2N$ for discrimination. Before conducting group discrimination, \change{in the ``aggregation'' phase}, all data samples are aggregated with the same aggregation technique, e.g., sum-, mean-, and linear aggregation.

\noindent \textbf{Group Discrimination.} In the group discrimination process, we adopt a very simple binary cross entropy (BCE) loss to discriminate two groups of node \change{samples} as shown in Equation \ref{eq:binary DGI}.
In our implementation, $ y_i$ is 0 and 1 for node embeddings in negative and positive groups. During model training, the model is optimised by categorising node embeddings in the collection of data samples into their corresponding class correctly. The loss is computed by comparing the \change{prediction} of a node $i$, i.e., a scalar, with its indicator $ y_i$. With the ease of BCE loss computation, the training process of \ourmethod\ is very fast and memory efficient.

\subsection{Model Inference}
During training, the model is optimised via loss minimisation with Equation \ref{eq:binary DGI}. The time complexity analysis of \ourmethod\ is provided in Appendix \ref{sec:complexity analysis}. \change{In the inference phase, we freeze the trained GNN encoder $g_{\theta}$ and obtain node embeddings $ \textbf{H}_{\theta}$ with the input $\mathcal{G}$.} 

Inspired by MVGRL~\cite{hassani2020contrastive}, which strengthens the output embeddings by including additional global information, we adopt a conceptually similar embedding reinforcement approach. Specifically, they obtain the final embeddings by summing up embeddings from two views: the original view comprising local information and the diffused view with global information. This operation reinforces the final embeddings and leads to model performance improvement. Nonetheless,
graph diffusion impairs the scalability of a model \cite{zheng2021towards} and hence cannot be directly applied in our embedding generation process. To avoid the diffusion computation, we have come up with a workaround in the virtue of the power of a graph to extract global information. The power of a graph can extend the message passing scope of $ \textbf{H}_{\theta}$ to n-hop neighbourhood, which encodes global information from distant neighbours. It can be formulated as follows: 
\vspace{-1mm}
\begin{equation}
     \textbf{H}_{\theta}^{global} = \textbf{A}^n\textbf{H}_{\theta},
    \label{graph power}
\end{equation}
where $ \textbf{H}_{\theta}^{global}$ is the global embedding, and $ \textbf{A}$ is the adjacency matrix of the graph $\mathcal{G}$. 
It is notable that this operation can be easily decomposed with the associative property of matrix multiplication and is easy to compute.
To show the easiness of such computation, we conduct an experiment showing its time consumption on various datasets in Appendix \ref{sec:graph power}. Finally, the final embedding can be achieved by $ \textbf{H} =  \textbf{H}_{\theta}^{global} + \textbf{H}_{\theta}$, 
which can be used for downstream tasks. \change{In our experiment, we conduct node classification tasks. Following the common practice of GCL methods \cite{velivckovic2018deep,zhu2020deep,thakoor2021bootstrapped,peng2020graph, zbontar2021barlow}, these tasks are performed by using the final embeddings $\textbf{H}$ to train and test a simple logistic regression classifier.}

%% file: 6_Related_Work.tex
\noindent\textbf{Graph Neural Networks (GNNs).}
are generalised deep neural networks for  graph-structured data. GNNs mainly have two categories, spectral-based GNNs and spatial-based GNNs. Spectral GNNs attempt to use eigen-decomposition to obtain the spectral-based representation of graphs, whereas spatial GNNs focus on using spatial neighbours of nodes for message passing. 
Extending spectral-based methods to the spatial domain, GCN~\cite{kipf2016semi} utilises first-order Chebyshev polynomial filters to approximate spectral-based graph convolution. Taking the weight of spatial neighbours in consideration, 
GAT~\cite{velivckovic2017graph}, improves GCN by introducing attention module in message passing. To decouple message passing from neural networks, SGC~\cite{wu2019simplifying} simplifies GCN by removing non-linearity and weight matrices in graph convolution layers. 
However, these studies cannot handle datasets with limited or no labels. Graph contrastive learning has been recently exploited to address this issue.

\noindent\textbf{Graph Contrastive Learning (GCL).} aims to alleviate the reliance on labelling information in model training based on the concept of mutual information (MI). Specifically,  GCL approaches maximise MI between instances with similar semantic information, and minimise MI between dissimilar instances. For example, 
DGI~\cite{velivckovic2018deep} 
builds contrastiveness between node embeddings and a summary vector (i.e., a graph level embedding obtained by averaging all node embeddings) with a JSD estimator. To improve DGI, MVGRL~\cite{hassani2020contrastive} and GMI~\cite{peng2020graph} extends the idea of DGI by introducing multi-view contrastiveness with diffusion augmentation, and focusing on a local scope with the first-order neighbourhood, respectively. Adopting InfoNCE loss, GRACE~\cite{zhu2020deep} applies augmentation techniques 
to create two augmented views and inject contrastiveness between them. 
Though these GCL methods have successfully outperformed some supervised baselines in benchmark datasets, these methods suffer from significant limitations, including \change{time-consuming training, memory inefficiency,} and poor scalability. 
In contrast, \ourmethod\ \change{requires much less time in training and posses high scalability.}



\noindent\textbf{Scalable GNNs. }
Efficiency is  a bottleneck for most existing GNNs to handle large graphs. To address this challenge, there are mainly three categories of approaches: layer-wise sampling (e.g., GraphSage~\cite{hamilton2017inductive}), graph sampling methods such as Cluster-GCN~ \cite{chiang2019cluster} and GraphSAINT ~\cite{klicpera2018predict}, and linear models, e.g., SGC~\cite{wu2019simplifying} and PPRGo \cite{bojchevski2020scaling}. GraphSage~\cite{hamilton2017inductive} introduces a neighbour-sampling approach, which creates fixed-size subgraphs for each node. Underpinned by graph sampling, Cluster-GCN \cite{chiang2019cluster} decomposes a large-scale graph into multiple subgraphs based on clustering, while GraphSAINT~ \cite{klicpera2018predict} utilises light-weight graph samplers along with a normalisation technique for biases elimination in mini-batches. Linear models, SGC~\cite{wu2019simplifying} and PPRGo~ \cite{bojchevski2020scaling}, decouple graph convolution from embedding transformation (i.e., matrix multiplication with weight matrices), and leverage Personalised PageRank to encode multi-hop neighbourhood, respectively. However, all these methods only focus on supervised learning on graphs. For unsupervised/self-supervised learning settings where no labelled supervision signal is available, 
these frameworks are not applicable.  
The \change{closest} works to ours to handle large scale graph datasets under self-supervised settings are BGRL~\cite{thakoor2021bootstrapped} and GBT~\cite{bielak2021graph}. \change{They  try to improve the contrastive losses by removing negative samples. However, BGRL~\cite{thakoor2021bootstrapped} and GBT~\cite{bielak2021graph} still require much more time in training compared with \ourmethod.}




%% file: 7_Experiment.tex
\begin{wraptable}{r}{0.49\textwidth}
\vspace{-1.85cm}
\footnotesize
	\caption{Model performance of node classification on 5 datasets. \textbf{X, A} and \textbf{Y} represent feature, adjacency matrix, and labels. Best performance for each dataset is in \textbf{bold}. Comp and Photo refer to Amazon Computers and Amazon Photos.}
	\resizebox{0.5\columnwidth}{!}{
	\begin{tabular*}{0.7\textwidth}{l|p{1.0cm}<{\centering}|p{0.9cm}<{\centering}p{0.9cm}<{\centering}p{0.9cm}<{\centering}p{0.9cm}<{\centering}p{0.9cm}<{\centering}}
		\toprule
		\textbf{Data} & \textbf{Method} & \textbf{Cora} & \textbf{CiteSeer} & \textbf{PubMed} & \textbf{Comp} & \textbf{Photo} \\
		\midrule
		\textbf{X, A, Y} & GCN & 81.5 & 70.3 & 79.0 & 76.3$\pm{0.5}$  & 87.3$\pm{1.0}$  \\
		\textbf{X, A, Y} & GAT & 83.0$\pm{0.7}$ & 72.5$\pm{0.7}$ & 79.0$\pm{0.3}$  & 79.3$\pm{1.1}$  & 86.2$\pm{1.5}$ \\
		\textbf{X, A, Y} & SGC & 81.0$\pm{0.0}$ & 71.9$\pm{0.1}$ & 78.9$\pm{0.0}$ & 74.4$\pm{0.1}$  & 86.4$\pm{0.0}$\\
		\textbf{X, A, Y} & CG3 & 83.4$\pm{0.7}$ & \textbf{73.6$\pm{0.8}$} & 80.2$\pm{0.8}$ & 79.9$\pm{0.6}$  & 89.4$\pm{0.5}$ \\
		\midrule
			\textbf{X, A}	& DGI & 81.7$\pm{0.6}$ & 71.5$\pm{0.7}$ & 77.3$\pm{0.6}$  &75.9$\pm{0.6}$& 83.1$\pm{0.5}$ \\
			\textbf{X, A}	& GMI & 82.7$\pm{0.2}$ & 73.0$\pm{0.3}$ & 80.1$\pm{0.2}$ & 76.8$\pm{0.1}$ & 85.1$\pm{0.1}$ \\
			\textbf{X, A}	& MVGRL & 82.9$\pm{0.7}$ & 72.6$\pm{0.7}$ & 79.4$\pm{0.3}$  & 79.0$\pm{0.6}$ & 87.3$\pm{0.3}$\\
			\textbf{X, A}	& GRACE & 80.0$\pm{0.4}$ & 71.7$\pm{0.6}$ & 79.5$\pm{1.1}$ & 71.8$\pm{0.4}$  & 81.8$\pm{1.0}$ \\
			\change{\textbf{X, A}}	& \change{GraphCL} & \change{82.5$\pm{0.2}$} & \change{72.8$\pm{0.3}$} & \change{77.5$\pm{0.2}$}  & \change{OOM} & \change{79.5$\pm{0.4}$} \\
			\textbf{X, A}	& BGRL & 80.5$\pm{1.0}$ & 71.0$\pm{1.2}$ &  79.5$\pm{0.6}$ & 89.2$\pm{0.9}$ & 91.2$\pm{0.8}$ \\
			\textbf{X, A}	& GBT & 81.0$\pm{0.5}$ & 70.8$\pm{0.2}$ & 79.0$\pm{0.1}$ & 88.5$\pm{1.0}$ & 91.1$\pm{0.7}$ \\
		\midrule
			\textbf{X, A}	& \textbf{\ourmethod} & \textbf{83.9}$\pm{0.4}$ & 73.0$\pm{0.6}$& \textbf{81.3}$\pm{0.8}$ &  \textbf{90.1}$\pm{0.9}$  & \textbf{92.5}$\pm{0.6}$ \\
		\bottomrule
	\end{tabular*}}
\label{tab: classification results}
\vspace{-0.4cm}
\end{wraptable}

We evaluate the effectiveness of our model using eight benchmark datasets of different sizes. These datasets include five small- and medium-scale datasets: Cora, CiteSeer, PubMed \cite{sen2008collective}, Amazon Computers, and Amazon Photos \cite{shchur2018pitfalls}, as well as large-scale datasets ogbn-arxiv, ogbn-products and ogbn-papers100M. Notably, ogbn-papers100M is the largest dataset provided by Open Graph Benchmark\cite{hu2020open} for node property prediction tasks. It has over 110 million nodes and 1 billion edges. The statistics of these datasets are summarised in Appendix \ref{sec:dataset stat}. To ensure reproducibility, \textbf{the detailed experiment settings and computing infrastructure are summarised in Appendix \ref{sec:exp setting}. } The source code is already open sourced\footnote{https://github.com/zyzisastudyreallyhardguy/Graph-Group-Discrimination}.

\begin{wraptable}{r}{0.5\textwidth}
\vspace{-0.5cm}
\footnotesize
  \caption{Comparison of training time per epoch in \underline{seconds} between six GCL-based methods and \ourmethod\ on five datasets. Improve means how many times are \ourmethod\ faster than baselines. `-' means the improvement range.}
  \label{tab:time}\vspace{-2mm}
  \resizebox{0.5\columnwidth}{!}{
  \begin{tabularx}{1.55\linewidth}{lp{1.5cm}<{\centering} p{1.5cm}<{\centering}p{1.5cm}<{\centering}p{1.5cm}<{\centering}p{1.5cm}<{\centering}}
    \toprule
    \textbf{Method} &\textbf{Cora} & \textbf{CiteSeer} &\textbf{PubMed} &\textbf{Comp} &\textbf{Photo}\\
    \midrule
    DGI & 0.085 & 0.134 & 0.158 & 0.171 & 0.059\\
    GMI & 0.394 & 0.497 & 2.285 & 1.297 & 0.637\\
    MVGRL & 0.123 & 0.171 & 0.488 & 0.663 & 0.468\\
    GRACE & 0.056 & 0.092 & 0.893 & 0.546 & 0.203\\
    \change{GraphCL} & \change{0.073} & \change{0.085} & \change{0.123} & \change{OOM} & \change{0.188}\\
    BGRL & 0.085 & 0.094 & 0.147 & 0.337 & 0.273\\
    GBT & 0.073 & 0.072 & 0.103 & 0.492 & 0.173\\
    \midrule
    \ourmethod & 0.010 & 0.021 & 0.015 & 0.016 & 0.009\\
    \midrule
    Improve & 7.3-39.4$\times$ & 3.4-23.7$\times$ & 6.9-152.3$\times$ & 10.7-15.3$\times$ & 19.2-70.8$\times$ \\
  \bottomrule
  \label{training time}
\end{tabularx}}
\vspace{-5mm}
\end{wraptable}

\subsection{Evaluating on Small- and Medium-scale Datasets}

\label{sec:evalsmall}
We compare \ourmethod\ with ten baselines including four supervised GNNs (i.e., GCN~\cite{kipf2016semi}, GAT~\cite{velivckovic2017graph}, SGC~\cite{wu2019simplifying}, and CG3~\cite{wan2020contrastive}) and six GCL methods (i.e., DGI~\cite{velivckovic2018deep}, GMI~\cite{peng2020graph}, MVGRL~\cite{hassani2020contrastive}, GRACE~\cite{zhu2020deep}, BGRL~\cite{thakoor2021bootstrapped} and GBT~\cite{bielak2021graph}) on five small- and medium scale benchmark datasets. In the experiment, we follow the same data splits as ~\cite{yang2016revisiting} for Cora, CiteSeer and PubMed. For Amazon Computers and Photos, we use a random split setting, which randomly allocates 10/10/80\% of data to training/validation/test set, respectively.
The model performance is measured using the averaged classification accuracy with five results along with standard deviations and reported in Table \ref{tab: classification results}.


\begin{wraptable}{r}{0.5\textwidth}
\vspace{-4mm}
\footnotesize
  \caption{Comparison of memory consumption in \underline{MBs} of six GCL baselines and \ourmethod\ on five datasets.}\vspace{-2mm}
  \label{tab:memory}
  \resizebox{0.5\columnwidth}{!}{
  \begin{tabularx}{1.55\linewidth}{lp{1.5cm}<{\centering} p{1.5cm}<{\centering}p{1.5cm}<{\centering}p{1.5cm}<{\centering}p{1.5cm}<{\centering}}
    \toprule
    \textbf{Method} &\textbf{Cora} & \textbf{CiteSeer} &\textbf{PubMed} &\textbf{Comp} &\textbf{Photo}\\
    \midrule
    DGI & 4,189 & 8,199 & 11,471 & 7,991 & 4,946 \\
    GMI & 4,527 & 5,467 & 14,697 & 10,655 & 5,219 \\
    MVGRL & 5,381 & 5,429 & 6,619 & 6,645 & 6,645\\
    GRACE & 1,913 & 2,043 & 12,597 & 8,129 & 4,881\\
    \change{GraphCL} & \change{4,163} & \change{8,249} & \change{11,555} & \change{OOM} & \change{9,083}\\
    BGRL & 1,627 & 1,749 & 2,299 & 5,069 & 3,303 \\
    GBT & 1,651 & 1,799 & 2,461 & 5,037 & 2,641\\
    \midrule
    \ourmethod &  1,475 & 1,587 & 1,629 & 1,787 & 1,637\\
    \midrule
    Improve &  10.7-72.6\% & 11.8-80.6\% & 27.2-85.8\% & 64.5-83.2\% & 38.0-75.4\%\\
  \bottomrule
\end{tabularx}
}
\vspace{-4mm}
\end{wraptable}
\noindent \textbf{Accuracy.} From Table \ref{tab: classification results}, we can observe that GGD generally outperforms all baselines
in all datasets. The only exception is on CiteSeer dataset, where the semi-supervised method, CG3\cite{wan2020contrastive}, slightly outperforms \ourmethod, which still provides the 2nd best performance. \change{In this experiment, we use the officially released code of GraphCL~\cite{you2020graph}, BGRL~\cite{thakoor2021bootstrapped} and GBT~\cite{bielak2021graph} to reproduce the result, while the other results are sourced from previous studies~\cite{wan2020contrastive, jin2021multi}.}

\noindent \textbf{Efficiency and Memory Consumption.}  \ourmethod\ is substantially more efficient than other self-supervised baselines in time and memory consumption as shown in Table \ref{tab:time} and Table \ref{tab:memory}. Remarkably, \ourmethod\ is \textbf{19.2 times faster} in Amazon Photos for training time per epoch, and consumes \textbf{64.5\% less memory} in Amazon Computers for memory consumption than the most efficient baseline (i.e., GBT~\cite{bielak2021graph}). The dramatic boost of time and memory efficiency of \ourmethod\ is contributed to the exclusion of similarity computation, which enables model training without multiplication of node embeddings. 

\subsection{Evaluating on Large-scale datasets}
\label{sec:large}
To evaluate the scalability of \ourmethod, we choose three large-scale datasets from Open Graph Benchmark~\cite{hu2020open}, which are ogbn-arxiv, ogbn-products, and ogbn-papers100M. ogbn-papers100M is the most challenging large-scale graph available in Open Graph Benchmark for node property prediction with over 1 billion edges and 110 million nodes. Extending to extremely large graphs (i.e., ogbn-products and ogbn-papers100M), we adopt a Neighbourhood Sampling strategy, which is described in Appendix \ref{sec:exp setting}.


\begin{wraptable}{r}{0.52\textwidth}
\vspace{-8mm}
\footnotesize
  \caption{Node classification result and efficiency comparison on ogbn-arxiv. 
  `epo' means epoch. `Time' means training time per epoch (in seconds). `Total' is total training time (Number of epochs $\times$ `Time'). OOM indicates out-of-memory on Nvidia A40 (48GB). Number after $\backslash$ means the hidden size of \ourmethod.}
  \label{tab:ogbn-arxiv}\vspace{-2mm}
  \resizebox{0.5\columnwidth}{!}{
  \begin{tabularx}{1.55\linewidth}{lp{0.9cm}<{\centering} p{1.2cm}<{\centering} |p{2.5cm}<{\centering} p{0.3cm}<{\centering}p{1.0cm}<{\centering}}
    \toprule
    \textbf{Method} &\textbf{Valid} & \textbf{Test} & \textbf{Memory} & \textbf{Time} & \textbf{Total}\\
    \midrule
    Supervised GCN & 73.0$\pm{0.2}$ & 71.7$\pm{0.3}$ & - & - & -  \\
    \midrule
    MLP & 57.7$\pm{0.4}$ & 55.5$\pm{0.2}$ & - & - & - \\
    Node2vec & 71.3$\pm{0.1}$ & 70.1$\pm{0.1}$ & - & - & - \\
    \midrule
    DGI & 71.3$\pm{0.1}$ & 70.3$\pm{0.2}$ & - & - & -\\
    GRACE(10k epos) & 72.6$\pm{0.2}$ & 71.5$\pm{0.1}$ & - & - & -\\
    BGRL(10k epos) & 72.5$\pm{0.1}$ & 71.6$\pm{0.1}$ & OOM (Full-graph) & / &/ \\
    GBT(300 epos) & 71.0$\pm{0.1}$ & 70.1$\pm{0.2}$ & 14,959MB & 6.47 &  1,941.00 \\
    \midrule
    \ourmethod (1 epo$\backslash$1500) & 72.7$\pm{0.3}$ & 71.6$\pm{0.5}$ & 14,666MB & 0.95 & 0.95|2,043$\times$ \\
    \ourmethod (1 epo$\backslash$256) & 71.0$\pm{0.2}$ & 70.3$\pm{0.3}$ & 4,513MB|69.8\% & 0.18 & 0.18|10,783$\times$ \\
  \bottomrule
\end{tabularx}}
\vspace{-3mm}
\end{wraptable}
\noindent \textbf{ogbn-arxiv \& ogbn-products.}
For ogbn-arxiv, we compare \ourmethod\ against four self-supervised baselines (i.e., DGI~\cite{velivckovic2018deep}, GRACE~\cite{zhu2020deep}, BGRL~\cite{thakoor2021bootstrapped}and GBT~\cite{bielak2021graph}), whereas BGRL~\cite{thakoor2021bootstrapped} and GBT~\cite{bielak2021graph} are selected to be compared for ogbn-products. In addition, we include the performance of MLP, Node2vec~\cite{grover2016node2vec}, and supervised GCN~\cite{kipf2016semi} sourced from~\cite{hu2020open} in Table \ref{tab:ogbn-arxiv} and Table \ref{tab:ogbn-products}. For memory and training time comparison, we only compare \ourmethod\ with the two most efficient baselines (i.e., BGRL and GBT according to Tables \ref{tab:time} and \ref{tab:memory}). In ogbn-arxiv, we reproduce BGRL~\cite{thakoor2021bootstrapped} and found it fails to process ogbn-arxiv in full batch. Thus, we only compare \ourmethod\ and GBT in this dataset, which can successfully train in full-graph processing mode.

\begin{wrapfigure}{r}{0.6\textwidth} 
\vspace{-2mm}
     \begin{subfigure}
         \centering
         \includegraphics[scale = 0.25]{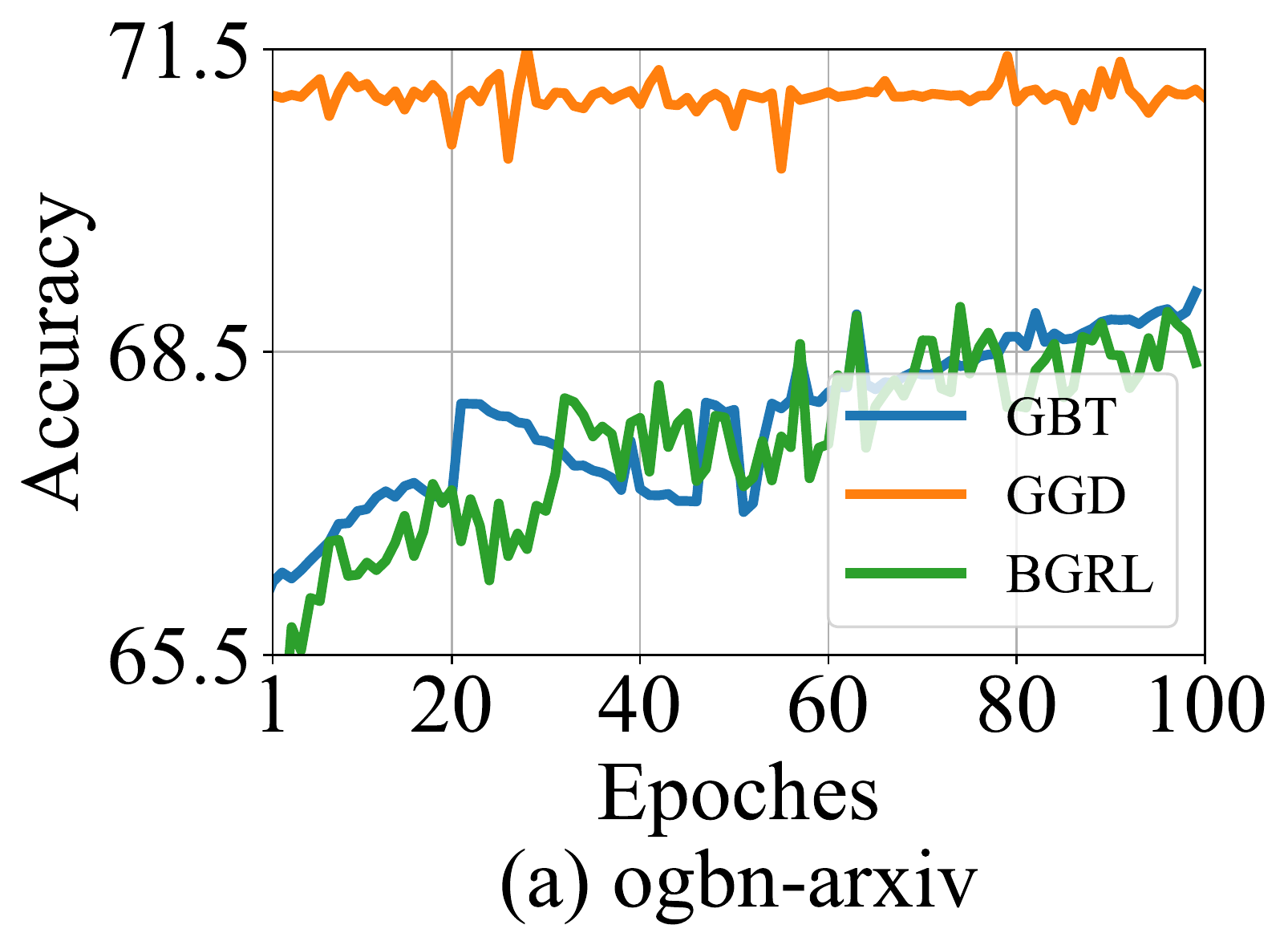}
         \label{fig:tsne-cora_1}
     \end{subfigure}
     \hspace{-5mm}
     \begin{subfigure}
         \centering
         \includegraphics[scale = 0.25]{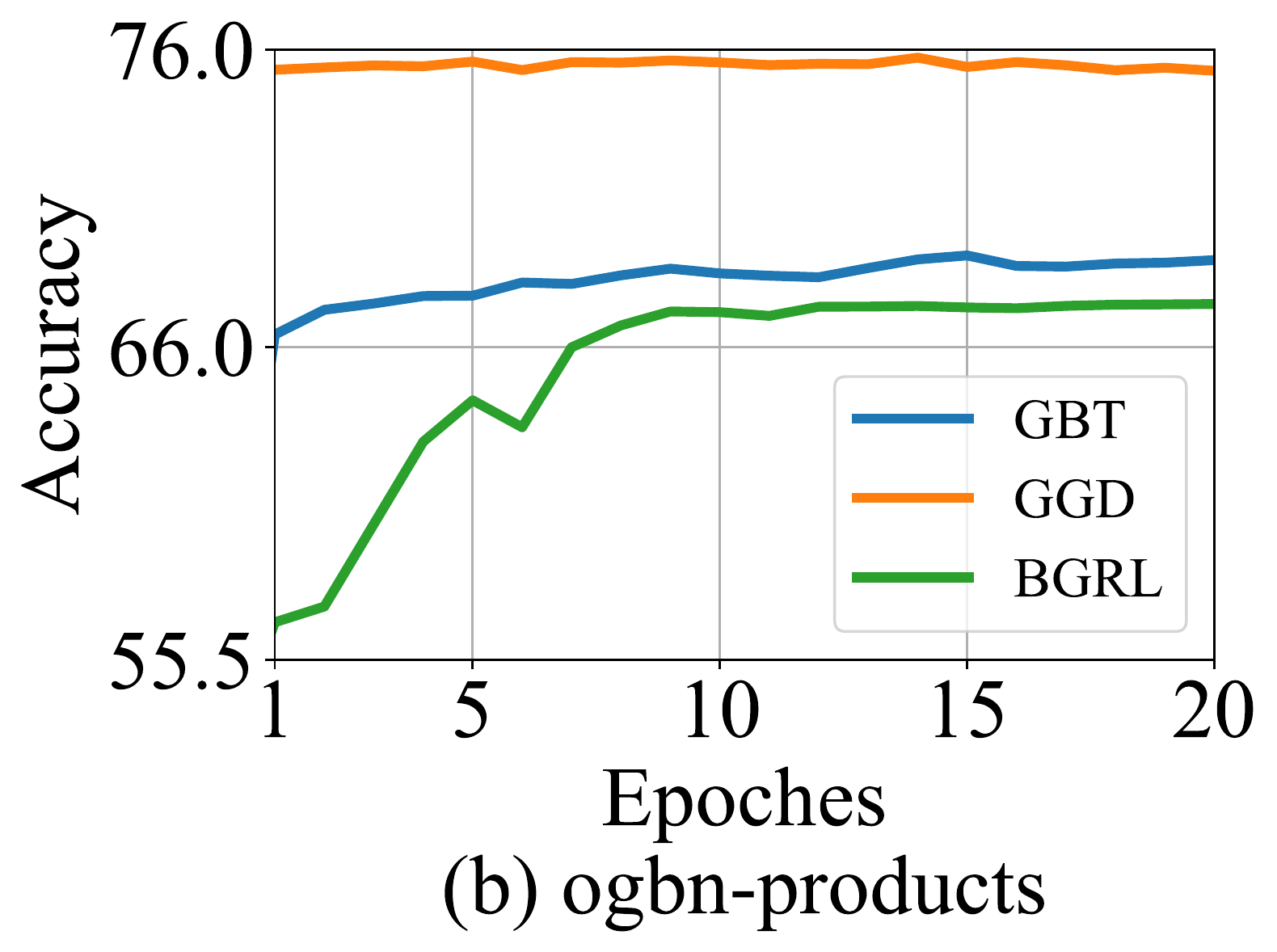}
         \label{fig:tsne-dgi_1}
     \end{subfigure}
     \vspace{-4mm}
     \caption{Convergence speed comparison among \ourmethod, BGRL\cite{thakoor2021bootstrapped} and GBT~\cite{bielak2021graph}. X-axis means number of epochs, while Y-axis represents the accuracy on test set.}
     \label{convergence}
    \vspace{-2mm}
\end{wrapfigure}

\begin{wraptable}{r}{0.5\textwidth}
\vspace{-3mm}
\footnotesize
  \caption{Node classification result and efficiency comparison on ogbn-products. }
  \label{tab:ogbn-products}\vspace{-2mm}
  \resizebox{0.5\columnwidth}{!}{
  \begin{tabularx}{1.55\linewidth}{lp{1.1cm}<{\centering} p{1.1cm}<{\centering} |p{1.6cm}<{\centering} p{0.8cm}<{\centering}p{1.1cm}<{\centering}}
    \toprule
    \textbf{Method} &\textbf{Valid} & \textbf{Test} & \textbf{Memory} & \textbf{Time} & \textbf{Total}\\
    \midrule
    Supervised GCN & 92.0$\pm{0.0}$ & 75.6$\pm{0.2}$ & - & - \\
    \midrule
    MLP & 75.5$\pm{0.0}$ & 61.1$\pm{0.0}$ & - & - \\
    Node2vec & 70.0$\pm{0.0}$ & 68.8$\pm{0.0}$ & - & - \\
    \midrule
    BGRL (100 epos) & 78.1$\pm{2.1}$ & 64.0$\pm{1.6}$ & 29,303MB & 53m16s & 5,326m40s \\
    GBT (100 epos) & 85.0$\pm{0.1}$ & 70.5$\pm{0.4}$ & 20,419MB & 48m38s & 4,863m20s\\
    \midrule
    \ourmethod (1 epo) & 90.9$\pm{0.5}$ & \textbf{75.7}$\pm{0.4}$ & 4,391MB|78.5\% & 12m46s & 12m46s|381$\times$\\
  \bottomrule
\end{tabularx}
}
\vspace{-2mm}
\end{wraptable}
From Table \ref{tab:ogbn-arxiv} and Table \ref{tab:ogbn-products}, we can see \ourmethod\ remarkably achieves the state-of-the-art performance using only one epoch to train. As a result, \ourmethod\ is 10,783 times faster than the most efficient baseline, i.e., GBT~\cite{bielak2021graph}, on total training time to reach the desirable performance in ogbn-arxiv. Please note that the number of epoches in our experiment is consistent with the optimal choice of this hyperparameter specified in GBT~\cite{bielak2021graph}. For ogbn-products, we are 381 $\times$ faster than GBT~\cite{bielak2021graph} on total training time. 
Notably, our performance is significantly higher than GCL baselines using 100 epochs (i.e., 6\% and 5.2\% improvement on GBT~\cite{bielak2021graph} in validation and test set, respectively) with only one epoch training in this dataset. In addition, we compare the convergence speed among \ourmethod , BGRL~\cite{thakoor2021bootstrapped} and GBT~\cite{bielak2021graph} on ogbn-arxiv and ogbn-products, which are shown in Figure \ref{convergence}. For ogbn-arxiv, BGRL~\cite{thakoor2021bootstrapped} is running using batched processing with neighbour sampling. This figure shows the preeminence of our \ourmethod\ in convergence speed as \ourmethod\ can be well-trained with only one epoch (i.e., reaching the peak model performance in the first epoch and staying stable with increased epochs). In contrast, the other two baselines require comparatively much more epochs to gradually improve their performance. Compared with GCL baselines, \ourmethod\ achieves much faster convergence via Group Discrimination. We conjecture this is because GD-based method focuses on the general edge distribution of graphs instead of node-specific information. Inversely, GCL methods can suffer from convergence inefficiency as they may be easily distracted from too-detailed node-specific information during training.

\noindent \textbf{ogbn-papers100M.}
We further compare \ourmethod\  with BGRL~\cite{thakoor2021bootstrapped} and GBT~\cite{bielak2021graph} on ogbn-papers100M, the largest OGB dataset with billion scale edges. Other self-supervised learning algorithms such as DGI~\cite{velivckovic2018deep} and GMI~\cite{peng2020graph} fail to scale to such a large graph with a reasonable batch size (i.e., 256). We only report the performance of each algorithm after a single epoch of training in Table \ref{ogbn-papers100m} due to the extreme scale of the dataset and the limitation of our available resources. From the table, we can observe that \ourmethod\ outperforms the two GCL counterparts, BGRL~\cite{thakoor2021bootstrapped} and GBT~\cite{bielak2021graph} in both accuracy and efficiency. Specifically, \ourmethod\ achieves 60.2 in accuracy while BGRL and GBT reach 59.3 and 58.9 in test set, respectively. With only one epoch, these two algorithms may not be well trained. However, training each epoch of these two requires over 1 day and if we would like to train them for 100 epochs, then we will need 100+ GPU days, which is prohibitively impractical for general practitioners. In contrast, \ourmethod\ can be trained in about 9 hours to achieve a good result for this dataset, which is more appealing in practice.


%% file: 10_Explore_GD.tex
In this section, we explore the corruption technique in \ourmethod\ and provide the theoretical analysis of group discrimination.

\subsection{Exploring Corruption}
\label{ap:why}

\begin{wrapfigure}{r}{0.53\textwidth}
\vspace{-7mm}
    \centering
    \includegraphics{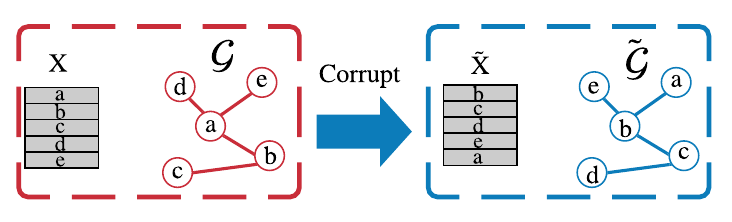}
    \caption{Corruption technique in DGI and MVGRL.}
    \label{fig:corruption}
    \vspace{-5mm}
\end{wrapfigure}

\begin{wraptable}{r}{0.5\textwidth}
\vspace{-0.5cm}
\footnotesize
  \caption{Node classification result and efficiency comparison on ogbn-papers100M. }
  \label{tab:ogbn-paper}\vspace{-2mm}
  \resizebox{0.48\columnwidth}{!}{
  \begin{tabularx}{1.35\linewidth}{lp{1.1cm}<{\centering} p{1.1cm}<{\centering}| p{1.77cm}<{\centering} p{1.0cm}<{\centering}}
    \toprule
    \textbf{Method} &\textbf{Validation} & \textbf{Test} &\textbf{Memory} &\textbf{Time}\\
    \midrule
    Supervised SGC & 63.3$\pm{0.2}$ & 66.5$\pm{0.2}$ & - & -\\
    \midrule
    MLP & 47.2$\pm{0.3}$ & 49.6$\pm{0.3}$ & - & -\\
    Node2vec & 55.6$\pm{0.0}$ & 58.1$\pm{0.0}$ & - & -\\
    \midrule
    BGRL (1 epoch)& 59.3$\pm{0.5}$ & 62.1$\pm{0.3}$ & 14,057MB & 26h28m\\
    GBT (1 epoch)& 58.9$\pm{0.4}$ & 61.5$\pm{0.5}$ & 13,185MB & 24h38m \\
    \midrule
    \ourmethod (1 epoch) & 60.2$\pm{0.3}$ & 63.5$\pm{0.5}$ & 4,105MB|68.9\% & 9h15m|2.7$\times$ \\
  \bottomrule
  \label{ogbn-papers100m}
\end{tabularx}}
\vspace{-8mm}
\end{wraptable}
Firstly, we explore the corruption technique used in DGI~\cite{velivckovic2018deep} and MVGRL~\cite{hassani2020contrastive}, which is shown in Figure \ref{fig:corruption}. These two studies corrupt the topology of a given graph $\mathcal{G}$ by shuffling the feature matrix $ \textbf{X}$. This is because by changing the node order of $ \textbf{X}$, the neighbouring structure of $\mathcal{G}$ is completely changed, e.g., neighbours of node $a$ become node $b$ neighbours. 

With the corruption technique, negative samples in the negative group are generated with incorrect edges. Thus, by discriminating the positive group (i.e., nodes generated with ground truth edges) and the negative group, we conjecture the model can distil valuable signals \change{by} learning how to identify nodes generated with correct topology and output effective node embeddings. To provide explanation to this, we present the theoretical analysis of group discrimination in the following section.

\subsubsection{Theoretical Analysis of Group Discrimination}
\label{ap:theory}
Group discrimination is learning to avoid making `mistakes' (i.e., bias the encoder towards avoiding mistaken samples). To explain this point, we first present Theorem~\ref{th: spearate} and then provide an intuitive explanation for group discrimination. 
\begin{theorem}
\label{th: spearate}
Given a graph $\mathcal{G}$, a corrupted graph $\tilde{\mathcal{G}}$, and a encoding network $g(\cdot)$, we consider the distribution of positive embeddings $g(\mathcal{G})$ as $P_{pos}$ and negative embeddings $g(\tilde{\mathcal{G}})$ as $P_{neg}$. Optimising the group discrimination loss is equivalent to maximising the Jensen-Shannon divergence between $P_{pos}$ and $P_{neg}$.
\end{theorem}
\begin{wrapfigure}{r}{0.65\textwidth}
\vspace{-2mm}
    \centering
    \includegraphics[scale = 0.55]{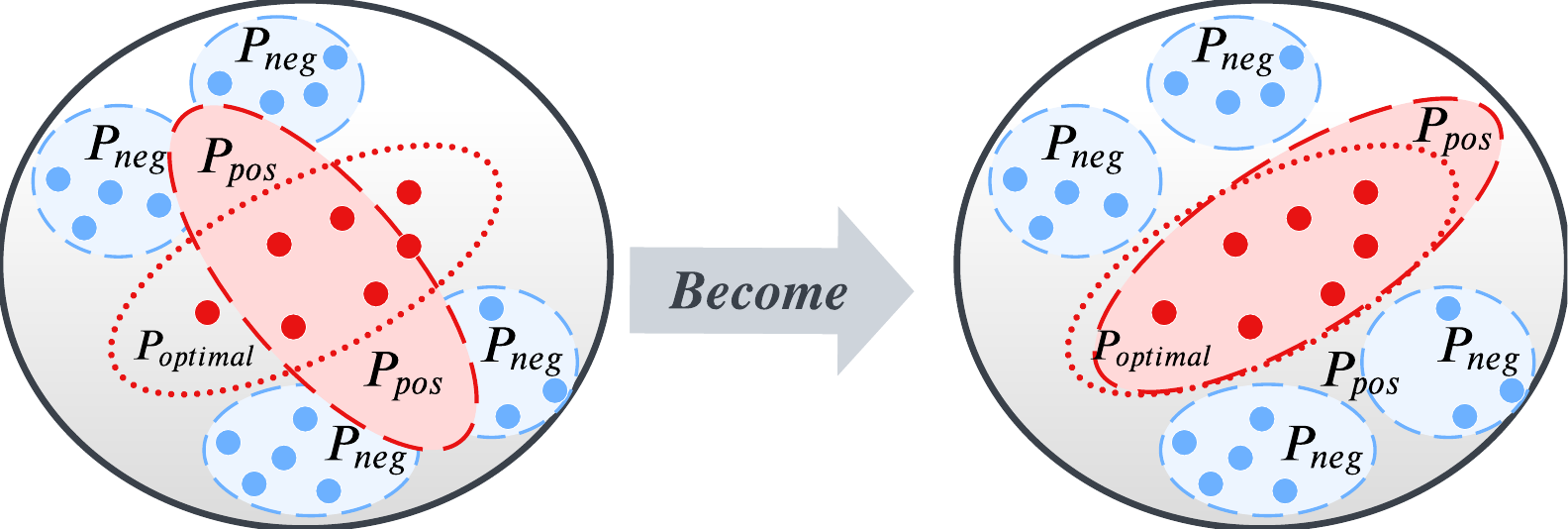}
    \caption{$P_{optimal}$ is the optimal distribution for node embeddings, $P_{pos}$ is the distribution of positive samples, $P_{neg}$ is the distribution of negative samples, blue nodes represent negative samples, and red nodes are samples in the optimal distribution. At the beginning, $P_{pos}$ is overlapped with $P_{neg}$. Then, $P_{pos}$ is gradually separated from $P_{neg}$ and ideally become closer to $P_{optimal}$.}
    \label{fig:neg}
\end{wrapfigure}
The proof for Theorem \ref{th: spearate} is presented in Appendix \ref{ap:proof_th2}. From the theorem above, we can see maximising the group discrimination loss $\mathcal{L}$ is the same as maximising $JS(P_{pos} \parallel P_{neg})$, where $JS$ represents the Jenson-Shannon divergence. Thus, by optimising the loss $\mathcal{L}$, $P_{pos}$ and $P_{neg}$ tend to be separated. \change{As a result, group discrimination is intuitively learning to avoid making `mistakes' (i.e., bias the encoder towards avoiding mistaken samples) as shown in Figure \ref{fig:neg}. This is because by separating  $P_{pos}$ and $P_{neg}$, $P_{pos}$ can gradually become similar to $P_{optimal}$, the optimal distribution for node embeddings. As $P_{optimal}$, is disjoint with $P_{neg}$, if the generated embeddings can avoid being similar to out-of-distribution samples, i.e., negative samples, it can be ideally closer to $P_{optimal}$. Therefore, the trained model can improve the quality of generated node embeddings for node samples.}

%% file: 8_Future_work.tex
In this paper, we have introduced a new self-supervised GRL paradigm: Group Discrimination, which achieves the same level of performance as GCL methods with much less resource consumption (i.e., training time and memory).
Some limitations of this work are we still have not explored some questions for GD. For example, can we extend the current binary Group Discrimination scheme (i.e., classifying \change{nodes generated with different topology}) to discrimination among multiple groups? Are there any other corruption technique to create a more difficult negative group for discrimination? 
More importantly, with the extremely efficient property, GD has the potential to be deployed to various real-world applications, e.g., recommendation systems, which have limited labelling information and desire fast computation with limited resources.

\begin{ack}
This research was partially supported by an Australian Research Council (ARC) Future Fellowship (FT210100097).

This work is supported in part by NSF under grants III-1763325, III-1909323,  III-2106758, and SaTC-1930941. 
\end{ack}

%% file: 9_checklist.tex

\begin{enumerate}

\item For all authors...
\begin{enumerate}
  \item Do the main claims made in the abstract and introduction accurately reflect the paper's contributions and scope?
    \answerYes{See Abstract and section~\ref{sec:intro}.}
  \item Did you describe the limitations of your work?
    \answerYes{See section \ref{sec: future work}}
  \item Did you discuss any potential negative societal impacts of your work?
    \answerNA{}
  \item Have you read the ethics review guidelines and ensured that your paper conforms to them?
    \answerYes{}
\end{enumerate}

\item If you are including theoretical results...
\begin{enumerate}
  \item Did you state the full set of assumptions of all theoretical results?
    \answerYes{Please refer to section \ref{sec:rethink} and Appendix \ref{proof 1}}
        \item Did you include complete proofs of all theoretical results?
    \answerYes{Please refer to section \ref{sec:rethink} and Appendix \ref{proof 1}}
\end{enumerate}

\item If you ran experiments...
\begin{enumerate}
  \item Did you include the code, data, and instructions needed to reproduce the main experimental results (either in the supplemental material or as a URL)?
    \answerYes{See Appendix~\ref{sec:exp setting} and the source code in supplemental material.}
  \item Did you specify all the training details (e.g., data splits, hyperparameters, how they were chosen)?
    \answerYes{See section \ref{sec:experiment} and Appendix~\ref{sec:exp setting}.}
        \item Did you report error bars (e.g., with respect to the random seed after running experiments multiple times)?
    \answerNo{}
        \item Did you include the total amount of compute and the type of resources used (e.g., type of GPUs, internal cluster, or cloud provider)?
    \answerYes{See section \ref{sec:experiment} and Appendix~\ref{sec:exp setting}.}
\end{enumerate}

\item If you are using existing assets (e.g., code, data, models) or curating/releasing new assets...
\begin{enumerate}
  \item If your work uses existing assets, did you cite the creators?
    \answerYes{See section \ref{sec:experiment} and Appendix~\ref{sec:exp setting}.}
  \item Did you mention the license of the assets?
    \answerYes{See section \ref{sec:experiment} and Appendix~\ref{sec:exp setting}.}
  \item Did you include any new assets either in the supplemental material or as a URL?
    \answerNo{We do not use any new datasets in this paper.}
  \item Did you discuss whether and how consent was obtained from people whose data you're using/curating?
    \answerNo{The datasets used in this paper are publicly available.}
  \item Did you discuss whether the data you are using/curating contains personally identifiable information or offensive content?
    \answerNA{}
\end{enumerate}

\item If you used crowdsourcing or conducted research with human subjects...
\begin{enumerate}
  \item Did you include the full text of instructions given to participants and screenshots, if applicable?
    \answerNA{}
  \item Did you describe any potential participant risks, with links to Institutional Review Board (IRB) approvals, if applicable?
    \answerNA{}
  \item Did you include the estimated hourly wage paid to participants and the total amount spent on participant compensation?
    \answerNA{}
\end{enumerate}

\end{enumerate}

%% file: Appendix_A.tex
\section{Appendix A}

\subsection{\change{Proof of Proposition 1}}
\label{proof 1}
\noindent\textit{Proof.} To prove Proposition \ref{th:th1}, given a graph $\mathcal{G} = \{\textbf{X},\textbf{A}\}$, where $\textbf{X} \in \mathbb{R}^{N \times D}$, $\textbf{A} \in \mathbb{R}^{N \times N}$ and a GNN encoder $g(\cdot)$. We consider $g(\cdot)$ as a one-layer GCN, which can be expressed as:
\begin{equation}
\label{eq: embed}
\begin{aligned}
    \textbf{H} &= \sigma(\hat{\textbf{D}}^{-\frac{1}{2}}\hat{\textbf{A}}\hat{\textbf{D}}^{-\frac{1}{2}}\textbf{ZW}),
\end{aligned}
\end{equation}
where $\sigma(\cdot)$ is non-linear activation function, $\textbf{H}  \in \mathbb{R}^{N \times D'}$ is the output embedding, $\textbf{Z} = norm(\textbf{X})$ ($norm(\cdot)$ is row normalisation), $\textbf{W} \in \mathbb{R}^{D \times D'}$ is learnable weight matrix, $\hat{\textbf{A}} = \textbf{A} + \textbf{I}$ (\textbf{I} is identity matrix), and $\hat{\textbf{D}}$ is the degree matrix for $\hat{\textbf{A}}$.

We consider $\tilde{\textbf{A}} = \hat{\textbf{D}}^{-\frac{1}{2}}\hat{\textbf{A}}\hat{\textbf{D}}^{-\frac{1}{2}}$ and $\textbf{V}=\tilde{\textbf{A}}\textbf{Z}$, where elements ${v}_{ik} \in \textbf{V}$, the $i$-th row and the $k$-th column element of $\textbf{V}$, are definite.
After that, we conduct $\textbf{P = VW}$, where $\textbf{P} \in \mathbb{R}^{N \times D'}$ and $\textbf{W}$ is Xavier initialised.

An elment $w_{mk} \sim U(-a, a)$ of $\textbf{W}$ is the $m$-th row and the $k$-th column of $\textbf{W}$, where $U(-a, a)$ is a uniform distribution. Here $a$ = $\alpha \times \sqrt{\frac{6}{D + D'}}$, where $D'$ is the hidden dimension, and $\alpha$ is a hyperparameter defaultly set to 1.
Then, we know the mean value of $w_{mk}$ distribution is 0 and its standard deviation is $\frac{2}{D + D'}$.


For any $p_{ik} \in \textbf{P}$, which is the $i$-th row and $k$-th column of $\textbf{P}$, it can be calculated as:
\begin{equation}
    p_{ik} = \sum^D_m v_{im}w_{mk},
\end{equation}
where $w_{mk} \in \textbf{W}$ is the $m$-th row and the $k$-th column of $\textbf{W}$, and $v_{im} \in \textbf{V}$ is the $i$-th row and $m$-th column of $\textbf{V}$.
According to the analysis on the weighted sum of uniform random variables \cite{kamgar1995distribution}, we have the mean $\mu$ and the standard deviation $\delta$ of $p_{ik}$:
\begin{equation}
    \mu = 0, \delta = \sqrt{\frac{2}{D+D'}\sum_{m=1}^{D}v_{im}^2}.
\end{equation}

Here we assume $p_{ik} \sim N(\mu,\delta^2)$ and use $[\mu-c \delta, \mu +c\delta ]$ to approximate the value scope of $p_{ik}$, where $c$ is a parameter that controls the precision of approximation. 

\textbf{Case 1 (ReLu/Leaky ReLu/PReLu).} In this case, we consider ReLu $\sigma_{ReLu}(\cdot)$, Leaky ReLu $\sigma_{LReLu}(\cdot)$, and PReLu $\sigma_{PReLu}(\cdot)$ as non-linear activation for the GNN encoder $g(\cdot)$. Here we take $h_{ik} = \sigma_{ReLu}(p_{ik})$ ($h_{ik} \in \textbf{H}$) as an example to show that the value range of $\sigma_{sig}(\sigma_{ReLu}(p_{ik}))$ is bounded with $\delta$. Please noted that, in practice, $\textbf{H}$ will be averaged to a summary vector $\textbf{s}$ and input to $\sigma_{sig}(\cdot)$. As $\sigma_{sig}(s)$ ($s \in \textbf{s}$) shares the same value range as $\sigma_{sig}(\sigma_{ReLu}(p_{ik}))$, for simplicity, we conducted analysis with $\sigma_{ReLu}(p_{ik})$ instead.

Given $p_{ik}$ as input, the value range of $\sigma_{ReLu}(p_{ik}) = max(0, p_{ik})$ is $[0, c\delta ]$. 
Similarly, we can also obtain $\sigma_{LReLu}(x) = max(0.01*x, x) \in [-0.01 c\delta,c\delta]$ and $\sigma_{PReLu}(x) = max(0, x) + b*min(0, x) \in [-bc\delta,c\delta]$(when $b\geq0$) or $[0, max(-bc\delta,c\delta)]$(when $b<0$).
Given $\sigma_{ReLu}(p_{ik}) \in [0, c\delta ]$, the infimum of $\sigma_{sig}(\sigma_{ReLu}(p_{ik}))$ is $\frac{1}{2}$ since sigmoid activation is a monotone increasing function.
To estimate the value range of final outputs, here we use the Taylor series of $\sigma_{sig}(\cdot)$ at 0 to estimate the upper bound \cite{daunizeau2017semi}, i.e.,
\begin{equation}
\label{eq:sig_taylor}
    \sigma_{sig}(x) \approx \sigma_{sig}(0) + \sigma^{\prime}_{sig}(0)x + \frac{1}{2}\sigma^{\prime\prime}_{sig}(0)x^2 + \cdot\cdot\cdot.
\end{equation}
Since $\sigma_{sig}(0)=\frac{1}{2}$, $\sigma_{sig}^{\prime}(x) =\sigma_{sig}(x)(1-\sigma_{sig}(x))$, $\sigma_{sig}^{\prime \prime}(x) = \sigma_{sig}(x)(1-\sigma_{sig}(x))(1-2 \sigma_{sig}(x))$, 
and $\frac{\partial^n}{\partial x^n} \sigma_{sig}(x) =\sigma_{sig}(x) \prod_{i=1}^n(1-i \sigma_{sig}(x))$, Equation (\ref{eq:sig_taylor}) can be expressed as:
\begin{equation}
\begin{aligned}
\label{eq:sig_value}
        \sigma_{sig}(x) \approx \sigma_{sig}(0) + \sigma^{\prime}_{sig}(0)x = \frac{1}{2} + \frac{1}{4}x.
\end{aligned}
\end{equation}
Thus, the value range of $\sigma_{sig}(\sigma_{ReLu}(p_{ik}))$ is $[\frac{1}{2},\frac{1}{2}+\frac{c\delta}{4}]$, which indicates $\sigma_{sig}(\sigma_{ReLu}(p_{ik})) \rightarrow \frac{1}{2}$ when $\delta \rightarrow 0$. 

\textbf{Case 2 (Sigmoid).} In this case, we consider Sigmoid $\sigma_{sig}(\cdot)$ as non-linear activation in $g(\cdot)$.  We can also obtain $\sigma_{sig}(p_{ik}) \in [\frac{1}{2}-\frac{c\delta}{4},\frac{1}{2}+\frac{c\delta}{4}]$ through Equation (\ref{eq:sig_value}) and $\sigma_{sig}(p_{ik}) \rightarrow \frac{1}{2}$ when $\delta \rightarrow 0$, which also indicates $\sigma_{sig}(\sigma_{sig}(p_{ik})) \rightarrow 0.62$ (i.e., $\sigma_{sig}(\frac{1}{2})$) when $\delta \rightarrow 0$.
In our experiments, we observe that $\delta$ owns a almost zero positive value, which results in our observations in Table \ref{tab:summ stat}.

\input{5_Exploring_GD}

\subsection{Evaluation on aggregation function}
\label{ap:aggregation}
\begin{wraptable}{r}{0.45\textwidth}
\vspace{-4mm}
\footnotesize
  \caption{The experiment result on three datasets with different aggregation function on node embeddings.}
  \label{tab:aggregation}
  \resizebox{0.45\columnwidth}{!}{
  \begin{tabularx}{1.4\linewidth}{lp{2.2cm}<{\centering} p{2.2cm}<{\centering} p{2.2cm}<{\centering}}
    \toprule
    \textbf{Method} &\textbf{Cora} & \textbf{CiteSeer} & \textbf{PubMed}\\
    \midrule
    Sum & 82.5 $\pm{0.2}$ & 71.7 $\pm{0.6}$ & 77.7 $\pm{0.5}$ \\
    Mean & 81.8 $\pm{0.5}$ & 71.8 $\pm{1.1}$ & 76.5 $\pm{1.2}$\\
    Min & 80.4 $\pm{1.3}$ & 61.7 $\pm{1.8}$& 70.1 $\pm{1.9}$\\
    Max & 71.4 $\pm{1.2}$& 65.3 $\pm{1.4}$ & 70.2 $\pm{2.8}$ \\
    linear & 82.2 $\pm{0.4}$ & 72.1 $\pm{0.7}$& 77.9 $\pm{0.5}$ \\
  \bottomrule
\end{tabularx}}
\vspace{-5mm}
\end{wraptable}
To explore the effect of other aggregation functions, we replace the summation function in Equation \ref{eq:DGI loss} with other aggregation methods, including mean-, minimum-, maximum- pooling and linear aggregation. We report the experiment results (i.e., averaged accuracy on five runs) in Table \ref{tab:aggregation}. The table shows that replacing the summation with other aggregation methods still works, while summation and linear aggregation achieve comparatively better performance. 

\subsection{Rethinking MVGRL}
\label{ap: rethink mvgrl}

\begin{figure}[htbp]
    \centering
    \includegraphics[scale = 0.8]{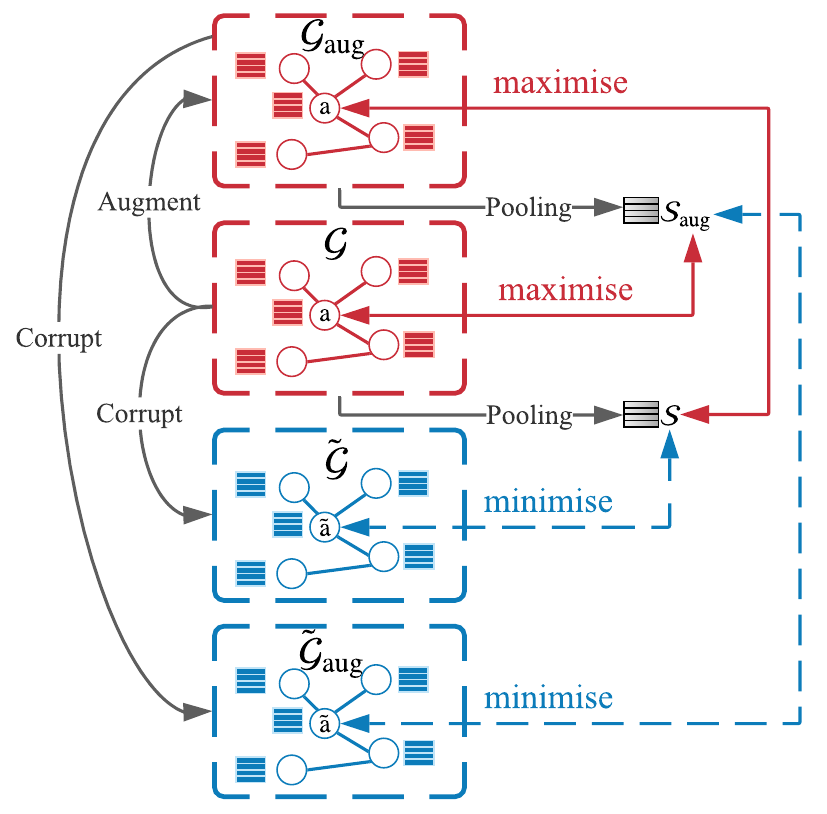}
    \caption{The architecture of MVGRL. Here augment means augmentation. $\textbf{s}$ is the summary vector based on $\mathcal{G}$, and $\textbf{s}_{aug}$ is the summary vector based on the augmented graph $\mathcal{G}_{aug}$.}
    \label{fig:mvgrl}
\end{figure}

Extending the architecture of DGI, MVGRL resorts to multi-view contrastiveness via additional augmentation.
Specifically, as shown in Figure \ref{fig:mvgrl}, it first uses the diffusion augmentation to create $\mathcal{G}_{aug}$. Then, it corrupts $\mathcal{G}$ and $\mathcal{G}_{aug}$ to generate negative samples $\tilde{\mathcal{G}}$ and $\tilde{\mathcal{G}}_{aug}$. To build contrastiveness, MVGRL also generates two summary vectors $\textbf{s}_{aug}$ and $\textbf{s}$ by averaging all embeddings in $\mathcal{G}_{aug}$ and $\mathcal{G}$, respectively. Based on the design of MVGRL, the model training is driven by mutual information maximisation between an anchor node embedding and its corresponding augmented summary vector. However, MVGRL has the same technical error in their official JSD-based implementation as DGI, which makes it also becomes a group-discrimination-based approach. 

Similar to Equation \ref{eq:binary DGI} of DGI, the proposed loss in MVGRL can also be rewritten as a binary cross entropy loss:
\begin{equation}
    \mathcal{L}_{MVGRL} = -\frac{1}{4N}(\sum_{i = 1}^{4N}y_i\log \hat{y}_i + (1 - y_i)\log(1 - \hat{y}_i)),
    \label{eq:mvgrl_loss}
\end{equation}
here the number of nodes is increased to $ 4N$ as we include both nodes in $\mathcal{G}_{aug}$ and $\tilde{\mathcal{G}}_{aug}$ as data samples. The indicator $y_i$ for $\mathcal{G}$ and $\mathcal{G}_{aug}$ are 1, while $\tilde{\mathcal{G}}$ and $\tilde{\mathcal{G}}_{aug}$ are considered as negative samples (i.e., the indicator $y_i$ for them are 0). To explore why MVGRL can achieve a better performance than DGI, we replace the original MVGRL loss with Equation \ref{eq:mvgrl_loss} and conduct ablation study of $ MVGRL_{bce}$ by removing different set of data samples in Equation \ref{eq:mvgrl_loss}, and report the experiment result in Table \ref{tab:mvgrl ablation}. From the table, the performance of $MVGRL_{bce}$ is on par with $ MVGRL$, which reconfirms the effectiveness of using the BCE loss. Also, we can observe that including $\mathcal{G}_{aug}$ and $\tilde{\mathcal{G}}_{aug}$ is the key of MVGRL surpassing DGI. With $\mathcal{G}_{aug}$ and $\tilde{\mathcal{G}}_{aug}$, the model performance of $ MVGRL_{bce} w/o \mathcal{G}_{aug}$ and $\tilde{\mathcal{G}}_{aug}$ is improved from 82.2 to 83.1. We conjecture this is because, with the diffusion augmentation, MVGRL is trained with the additional global information provided by the diffused view $\mathcal{G}_{aug}$. However, the diffusion augmentation involves expensive matrix inversion computation and significantly densifies the given graph, which requires much more memory and time to store and process than the original view. This can hinder the model from extending to large-scale datasets \cite{zheng2021towards}. 

\begin{table}
\footnotesize
\centering
  \caption{The ablation study of MVGRL from the perspective of Group Discrimination.}
  \label{tab:mvgrl ablation}
  \begin{tabularx}{0.72\linewidth}{lp{1.4cm}<{\centering} p{1.4cm}<{\centering} p{1.4cm}<{\centering}}
    \toprule
    \textbf{Method} &\textbf{Cora} & \textbf{CiteSeer} & \textbf{PubMed}\\
    \midrule
    $ MVGRL$ & 82.9 $\pm{0.9}$ & 72.6 $\pm{0.8}$ & 78.8 $\pm{0.6}$ \\
    $ MVGRL_{BCE}$ & 83.1 $\pm{0.6}$ & 72.8 $\pm{0.5}$ & 79.1 $\pm{1.1}$ \\
    $ MVGRL_{BCE}$ w/o $\mathcal{G}_{aug}$ & 81.2 $\pm{0.8}$ & 52.8 $\pm{3.1}$ & 76.6 $\pm{1.3}$ \\
    $ MVGRL_{BCE}$ w/o $\mathcal{G}$ & 82.1 $\pm{0.6}$ & 71.8 $\pm{1.1}$ & 77.1 $\pm{1.2}$\\
    $ MVGRL_{BCE}$ w/o $\tilde{\mathcal{G}}_{aug}$ & 81.1 $\pm{0.8}$ & 56.7 $\pm{2.1}$& 74.9 $\pm{1.3}$\\
    $ MVGRL_{BCE}$ w/o $\tilde{\mathcal{G}}$ & 82.7 $\pm{0.9}$ & 72.0 $\pm{0.9}$ & 78.6 $\pm{0.9}$\\
    $ MVGRL_{BCE}$ w/o $\mathcal{G}_{aug}$ and $\tilde{\mathcal{G}}_{aug}$ & 82.2 $\pm{0.6}$ & 71.8 $\pm{1.0}$ & 77.0 $\pm{0.8}$ \\
    $ MVGRL_{BCE}$ w/o $\mathcal{G}$ and  $\tilde{\mathcal{G}}$ & 83.1 $\pm{0.6}$ & 72.6 $\pm{0.6}$& 78.5 $\pm{1.4}$ \\
  \bottomrule
\end{tabularx}
\end{table}

\subsection{Complexity Analysis}
\label{sec:complexity analysis}
The time complexity of our method consists of two components: the \textit{siamese GNN} and the \textit{loss computation}.
Existing self-supervised baselines share similar time complexity for the first component. In \ourmethod, given a graph $\mathcal{G} = \{\textbf{X} \in \mathbb{R}^{N\times D}, \textbf{A} \in \mathbb{R}^{N\times N}\}$ in the sparse format, taking a $L$-layer GCN~\cite{kipf2016semi} encoder as an example, the time complexity is $ O(LND + LND^2)$ \footnote{Here we assume for simplicity that the graph is sparse with the number of edges $|E|=O(N)$.}. As we need to process both the augmented graph $\hat{\mathcal{G}}$ and the corrupted graph $\tilde{\mathcal{G}}$, \ourmethod\ requires the encoder computation twice. Then, the projector network (i.e., MLP) with $K$ linear layers will be applied to the encoder output, which takes $O(ND^2)$ for each layer in computation \footnote{For simplicity we assume the hidden dimension size is $D$. In practice, the hidden dimension size $D'$ will be smaller than $D$.}. Before group discrimination, we aggregate the generated embedding with aggregation techniques. Here we take the simple summation, which consumes $ O(ND)$ as example.
For the loss computation, we use the BCE loss, i.e., Equation \ref{eq:binary DGI}, to category summarised node embeddings, i.e., scalars. The time complexity of this final step is $ O(2N)$ (i.e., processing all data samples from the positive and negative groups). Ignoring the computation cost of the augmentation, \change{the overall time complexity of \ourmethod\ for computing a graph $\mathcal{G}$ is  $ O(2(LND + LND^2 + KND^2 + ND + N)) \rightarrow O\big(ND(L + LD + KD )\big)$, where we can see the time complexity is mainly contributed by the siamese GNN. Also, our model scales linearly w.r.t. the number of nodes $N$.}


\subsection{The power of Graphs}
\label{sec:graph power}
To show the easiness of computation for the power of graphs, we conduct an experiment to evaluate the time consumption for graph power computation on eight datasets, whose statistics are shown in Appendix \ref{sec:dataset stat}. Specifically, we set the hidden size of $ \textbf{H}_{\theta}$ to 256, and $ n$ is fixed to 10 for all datasets. The experiment results are shown as below:

\begin{table}[htp]
\footnotesize
\centering
  \caption{Graph power computation time in \underline{seconds} on eight benchmark datasets. The experiment is conducted using CPU: Intel Xeon Gold 5320. `Cite' , `Comp', `Photo', `Arxiv', `Products', `Papers' means Citeseer, Amazon Computer, Amazon Photo, ogbn-arxiv, ogbn-products and ogbn-papers100M.}
  \begin{tabularx}{0.97\linewidth}{lp{1.4cm}<{\centering} p{1.4cm}<{\centering}p{1.4cm}<{\centering}p{1.4cm}<{\centering}p{1.4cm}<{\centering}p{1.4cm}<{\centering}p{1.4cm}<{\centering}}
    \toprule
    \textbf{Cora} & \textbf{Cite} &\textbf{PubMed} &\textbf{Comp} &\textbf{Photo} &\textbf{Arxiv} &\textbf{Products} &  \textbf{Papers}\\
    \midrule
    5.4e-3& 7.3e-3 & 9.8e-3 & 1.2e-2 & 8.5e-3 & 2.2e-2 & 24.5 & 208.8 \\
  \bottomrule
\end{tabularx}
\vspace{-2mm}
\end{table}

This table shows that the computation of graph power is very trivial on small and medium size graphs, e.g., ogbn-arxiv, which has million of edges, consuming only 0.22 seconds. Extending to an extremely large graph, ogbn-papers100M, which has over 1 billion edges and 11 million nodes, the computation only requires 209 seconds (i.e., around three minutes), which is acceptable considering the sheer size of the dataset.

\subsection{Dataset Statistics}
\label{sec:dataset stat}
The following table presents the statistics of eight benchmark datasets including five small to medium -scale datasets and three large-scale datasets from OGB Graph Benchmark\cite{hu2020open}.
\begin{table}[H]
\caption{The statistics of eight benchmark datasets.}
\footnotesize
	\centering
	\begin{tabularx}{0.7\linewidth}{lp{1.3cm}<{\centering} p{1.0cm}<{\centering} p{1.3cm}<{\centering}p{1.3cm}<{\centering}}
		\toprule
		\textbf{Dataset} &  \textbf{Nodes} & \textbf{Edges} & \textbf{Features} & \textbf{Classes} \\ \midrule
		\textbf{Cora}              & 2,708    & 5,429        & 1,433            & 7          \\
		\textbf{CiteSeer}         & 3,327    & 4,732        & 3,703            & 6    \\
		\textbf{PubMed}           & 19,717   & 44,338      & 500               & 3           \\
		\textbf{Amazon Computers}     & 13,752     & 245,861       & 767       & 10        \\
		\textbf{Amazon Photo}    & 7,650     & 119,081       & 745              & 8              \\
		\textbf{ogbn-arxiv}    & 169,343     & 1,166,243       & 128             & 40      \\
		\textbf{ogbn-products}    & 2,449,029     & 61,859,140       & 100             & 47  \\
		\textbf{ogbn-papers-100M}    & 111,059,956     & 1,615,685,872       & 100             & 172  \\
 \bottomrule
	\end{tabularx}
	\label{tab:dataset}
\end{table}

\subsection{Experiment Settings \& Computing Infrastructure}
\label{sec:exp setting}
\textbf{Extending to Extremely Large Datasets.} Extending to extremely large graphs (i.e., ogbn-products and ogbn-papers100M), we adopt a simple neighbourhood sampling strategy introduced in GraphSage~\cite{hamilton2017inductive} to decouple model training from the sheer size of graphs. Specifically, we create a fixed size subgraph for each node, which is created by sampling a predefined number of neighbours in each convolution layer for sampled nodes. The same approach is employed in the testing phase to obtain final embeddings. 

\noindent\textbf{General Parameter Settings.}
In our experiment, we mainly tune four parameters for \ourmethod\ ,which are learning rate, hidden size, number of convolution layers in the GNN encoder, and number of linear layers in the projector. For simplicity, we set the the power of a graph for global embedding generation fixed to 5 for all datasets (i.e., Equation~\ref{graph power}). The parameter setting for each dataset is shown below:

\begin{table}[H]
\caption{Parameter settings on eight datasets. \change{`lr' means the learning rate for pretraining}, `num-conv' and `num-proj' represent number of convolution layers in GNNs and number of linear layers in projector, respectively. }
\footnotesize
	\centering
	\begin{tabularx}{0.7\linewidth}{lp{1.1cm}<{\centering} p{1.0cm}<{\centering} p{1.5cm}<{\centering}p{1.5cm}<{\centering}}
		\toprule
		\textbf{Dataset} &   \textbf{lr}  & \textbf{hidden} & \textbf{num-conv} & \textbf{num-proj} \\ \midrule
		\textbf{Cora}      & 1e-3    & 512  & 1            & 1         \\
		\textbf{CiteSeer} & 1e-5    & 1024      & 1           & 1    \\
		\textbf{PubMed}   & 1e-3   & 1024     & 1              & 1           \\
		\textbf{Amazon Computers}     & 1e-3   & 1024 & 1       & 1        \\
		\textbf{Amazon Photo}    & 1e-3    & 512 & 1              & 1              \\
		\textbf{ogbn-arxiv}    & 5e-5    & 1500 & 3           & 1      \\
		\textbf{ogbn-products}    & 1e-4    & 1024 & 4             & 4  \\
		\textbf{ogbn-papers-100M}    & 1e-3 & 256       & 3  & 1  \\
 \bottomrule
	\end{tabularx}
	\label{tab:parameter setting}
\end{table}

\noindent \textbf{Large-scale Datasets Parameter Settings.}
To decouple model training from the scale of graphs, we adopt the neighbouring sampling technique, which has three parameters: batch size, sample size, \change{and} number of hops to be sampled. Batch size refers to the number of nodes to be processed in one parameter optimisation step. Sample size means the number of nodes to be sampled in each convolution layer, and \change{the} number of hops determines the scope of the neighbourhood for sampling. In \ourmethod\ implementation, the batch size, sample size, and \change{the} number of hops are fixed to 2048, 12 and 3, respectively. 

\noindent \textbf{Memory and Training Time Comparison.} As memory and training time are very sensitive to hyper-parameters related to the structure of GNNs, including hidden size, number of convolution layers, and batch processing for large-scale datasets, e.g., batch size and number of neighbours sampled in each layer. Thus, in memory and training comparison, to be fair, we set all these parameters to be the same for all baselines and \ourmethod. The specific parameter setting for each dataset is shown below:

\begin{table}[H]
\vspace{-2mm}
\caption{Parameter settings on eight datasets for memory and training time comparison.}
\footnotesize
	\centering
	\begin{tabularx}{0.67\linewidth}{lp{1.0cm}<{\centering} p{1.2cm}<{\centering} p{1.0cm}<{\centering}p{1.5cm}<{\centering}}
		\toprule
		\textbf{Dataset} &  \textbf{hidden} & \textbf{num-conv} & \textbf{batch} & \textbf{num-neigh} \\ \midrule
		\textbf{Cora}      & 512    & 1  & -   & -          \\
		\textbf{CiteSeer} & 512  & 1     & -   & -   \\
		\textbf{PubMed}   & 256   & 1     & -     & -        \\
		\textbf{Amazon Computers}     & 256   & 1 & -   & -       \\
		\textbf{Amazon Photo}    & 256    & 1 & -   & -            \\
		\textbf{ogbn-arxiv}    & 256   & 3 &  - & -       \\
		\textbf{ogbn-products}    & 256    & 3 & 512   & 10        \\
		\textbf{ogbn-papers-100M}    & 128 & 3       & 512 & 10  \\
 \bottomrule
	\end{tabularx}
	\vspace{-2mm}
	\label{tab:parameter time&memory}
\end{table}

\noindent \textbf{Computing Infrastructure.}
For experiments in Section \ref{sec:rethink}, \ref{sec: method} and \ref{ap:why}, they are conducted using Nvidia GRID T4 (16GB memory) and Intel Xeon Platinum 8260 with 8 core. For experiments on large-scale datasets (i.e.,ogbn-arxiv, ogbn-products and ogbn-papers100M), we use NVIDIA A40 (48GB memory) and Intel Xeon Gold 5320 with 13 cores.

\subsection{\change{Algorithm}}
\change{We have summarised the overall procedure of \ourmethod \ in Algorithm \ref{alg} in \textbf{as follows}.
\input{algorithm}
Specifically, we first conduct augmentation on $\mathcal{G}$ to obtain $\hat{\mathcal{G}}$. Then, $\hat{\mathcal{G}}$ is corrupted to generate the corrupted graph $\tilde{{\mathcal{G}}}$ . In the encoding phase, we feed $\hat{\mathcal{G}}$ and $\tilde{{\mathcal{G}}}$ to GNN encoder $g_{\theta}$ and projector $f_{\theta}$ to generate embeddings $\rm{\hat{\textbf{H}}_{\theta}}$ for positive samples and $\rm{\tilde{\textbf{H}}_{\theta}}$ for negative samples. After that, we obtain the binary classification result by aggregating the concatenation of $\rm{\hat{\textbf{H}}_{\theta}}$ and $\rm{\tilde{\textbf{H}}_{\theta}}$. The result is a $2N$ dimension vector, which can be used for calculating the loss with a BCE loss. Finally, based on the calculated loss, trainable parameters in $g_{\theta}(\cdot)$ and $f_{\theta}(\cdot)$ can be updated. This training process will continue iteratively until we reach the predefined number of epochs $T$. 
When the training process is completed, we freeze the $g_{\theta}(\cdot)$ and $f_{\theta}(\cdot)$ and feed $\mathcal{G}$ to these two encoding components to obtain the local embedding, $\rm{\textbf{H}_{\theta}}$.  Then, we obtain global embedding $\textbf{H}_{\theta}^{global}$ with a global information injection operation. By summing   $\rm{\textbf{H}_{\theta}}$ and $\textbf{H}_{\theta}^{global}$, we can get the final embeddings $\textbf{H}$.
}

\change{
\subsection{Ablation Study}
In this section, we conduct an ablation study to evaluate the effectiveness of different components in \ourmethod. Specifically, we evaluate three variants of \ourmethod, including GGD$_{w/o\ aug}$, GGD$_{w/o\ proj}$, GGD$_{w/o\ power}$, which represent \ourmethod \ without augmentation, the projector and the global information injection process respectively. The experiment results on five small to medium size datasets are presented in Table \ref{tab: ablation study}. From the table, we can see without any mentioned component, the performance of \ourmethod\ degrades, which validates the effectiveness of these components.
It is worth noting that even without the global information injection in the inference phase, GGD$_{w/o\ power}$ still achieves the highest performance in 4 out of 5 datasets compared with six self-supervised baselines. This indicates that even without global information injection, \ourmethod\ is still effective.}

\begin{table}[h]
\footnotesize
\centering
  \caption{Ablation Study for GGD.}
  \resizebox{1.0\columnwidth}{!}{
  \begin{tabularx}{1.0\linewidth}{l|p{2.0cm}p{2.0cm}<{\centering}p{2.0cm}<{\centering} p{2.0cm}<{\centering}p{2.0cm}<{\centering}p{2.0cm}<{\centering}p{2.0cm}<{\centering}}
    \toprule
    \textbf{Method} &\textbf{Cora} & \textbf{CiteSeer} & \textbf{PubMed} & \textbf{Comp} & \textbf{Photo} \\
    \midrule
    GGD$_{w/o\ aug}$ & 83.6$\pm{0.3}$ & 72.4$\pm{0.4}$ & 81.2$\pm{0.2}$ & 89.6$\pm{0.4}$ & 92.2 $\pm{0.5}$\\
    GGD$_{w/o\ proj}$ & 83.0$\pm{0.5}$ & 72.5$\pm{0.4}$ & 81.1$\pm{0.4}$ & 89.4$\pm{0.5}$ &  91.6 $\pm{0.5}$\\
    GGD$_{w/o\ power}$ & 83.0$\pm{0.5}$ & 72.5$\pm{0.4}$ & 80.1$\pm{0.4}$ & 89.9$\pm{0.6}$ & 91.6$\pm{0.4}$ \\
    GGD & \textbf{83.9}$\pm{0.4}$ & \textbf{73.0}$\pm{0.6}$ & \textbf{81.3}$\pm{0.8}$ & \textbf{90.1}$\pm{0.9}$ & \textbf{92.5}$\pm{0.6}$\\
  \bottomrule
\end{tabularx}}
\label{tab: ablation study}
\end{table}

%% file: 5_Exploring_GD.tex
\subsection{Proof for Theorem \ref{th: spearate}}
\label{ap:proof_th2}

\noindent\textit{Proof.} The following proof is inspired by the theoretical proof in GAN \cite{goodfellow2014generative}. During training, we can rewrite the group discrimination loss in Equation \ref{eq:binary DGI} into the following objective (maximise to optimise):
\begin{equation}
\begin{aligned}
    \mathcal{L} &= \mathbb{E}_{\textbf{h} \sim P_{pos}} log(agg(\textbf{h})) + \mathbb{E}_{\textbf{h} \sim P_{neg}} log(1 - agg(\textbf{h})), \\ &= \int_\textbf{h} P_{pos}(\textbf{h}) log(agg(\textbf{h})) d\textbf{h} + \int_{\textbf{h}} P_{neg}(\textbf{h}) log(1 - agg(\textbf{h})) d\textbf{h}, 
\end{aligned}
\label{eq:rewrite}
\end{equation}
where $agg(\cdot)$ is the aggregation function to turn $\textbf{h}$ into the $1 \times 1$ prediction, $P_{pos}$ are the distribution of positive embeddings,  $P_{neg}$ are the distribution of negative embeddings. 
As our objective here is to maximise $\mathcal{L}$, and $P_{pos}(\textbf{h})>0; P_{neg}(\textbf{h}) > 0$, we can obtain the optimal solution for $agg(\textbf{h})$ is $\frac{P_{pos}(\textbf{h})}{P_{pos}(\textbf{h}) + P_{neg}(\textbf{h})}$. This is because for any $(a, b) \in \mathbb{R}^2 \backslash \{0, 0\}$, the maximum of a function $y = alog(x) + blog(1 - x)$ is achieved at $\frac{a}{a+b}$ \cite{goodfellow2014generative}. By replacing $agg(\textbf{h})$ with $\frac{P_{pos}(\textbf{h})}{P_{pos}(\textbf{h}) + P_{neg}(\textbf{h})}$ in Equation \ref{eq:rewrite}, we can obtain:
\begin{equation}
    \begin{aligned}
        \mathcal{L} &= \mathbb{E}_{\textbf{h }\sim P_{pos}} log(\frac{P_{pos}(\textbf{h})}{P_{pos}(\textbf{h}) + P_{neg}(\textbf{h})}) + \mathbb{E}_{\textbf{h} \sim P_{neg}} log(1 - \frac{P_{pos}(\textbf{h})}{P_{pos}(\textbf{h}) + P_{neg}(\textbf{h})}), \\&= \mathbb{E}_{\textbf{h} \sim P_{pos}} log(\frac{P_{pos}(\textbf{h})}{P_{pos}(\textbf{h}) + P_{neg}(\textbf{h})}) + \mathbb{E}_{\textbf{h} \sim P_{neg}} log(\frac{P_{neg}(\textbf{h})}{P_{pos}(\textbf{h}) + P_{neg}(\textbf{h})}).
    \end{aligned}
    \label{eq:loss js}
\end{equation}
From the equation above, we can see it looks similar to the Jensen-Shannon divergence between two distribution $P_1$ and $P_2$:
\begin{equation}
    JS(P_{1} \parallel P_{2}) = \frac{1}{2} \mathbb{E}_{\textbf{h} \sim P_{1}} log (\frac{\frac{P_{1}}{P_{1} + P_{2}}}{2}) + \frac{1}{2} \mathbb{E}_{\textbf{h} \sim P_{2}} log (\frac{\frac{P_{2}}{P_{1} + P_{2}}}{2}).
\end{equation}
Thus, we can rewrite Equation \ref{eq:loss js} as:
\begin{equation}
    \begin{aligned}
        \mathcal{L} &=  \mathbb{E}_{\textbf{h} \sim P_{pos}} log(\frac{\frac{P_{pos}(\textbf{h})}{P_{pos}(\textbf{h}) + P_{neg}(\textbf{h})}}{2}) + \mathbb{E}_{\textbf{h} \sim P_{neg}} log(\frac{\frac{P_{neg}(\textbf{h})}{P_{pos}(\textbf{h}) + P_{neg}(\textbf{h})}}{2}) -2log2, \\ &= 2 JS(P_{pos} \parallel P_{neg}) - 2log2,
    \end{aligned}
\end{equation}
\change{where we can see maximising $\mathcal{L}$ is the same as maximising $JS(P_{pos} \parallel P_{neg})$. Thus, by optimising $\mathcal{L}$, $P_{pos}$ and $P_{neg}$ tend to be separated.}

\change{\subsubsection{Connection with DGI}
\label{connection}
In this section, by building a connection with DGI, we explain our theoretical motivation for simplifying DGI objective to group discrimination loss and why group discrimination is efficient in computation time. We first present the Lemma \ref{le:lem1}, which is used in DGI \cite{velivckovic2018deep}: 
\begin{lemma}
\label{le:lem1}
Define $\normalfont{\{\textbf{H}^{g}\}^{|\textbf{H}|}_{g=1}}$ as a set of node embeddings drawn from distribution of graphs, $p(\normalfont{\textbf{H})}$, where $|\normalfont{\textbf{H}}|$ is finite number of elements, and $p(\normalfont{\textbf{H}^{g}) = p(\textbf{H}^{g'}}), \forall g, g'$. $\mathcal{R}(\cdot)$ is a deterministic readout function, which takes $\normalfont{\textbf{H}^g}$ as input and output the summary vector of the $g$-th graph, $\normalfont{\textbf{s}^g}$. $\normalfont{\textbf{s}^g}$ follows a marginal distributrion $p(\normalfont{\textbf{s}})$. Then, we assume $\mathcal{R}(\cdot)$ is injective and class balance, the upper bound of the error rate for the optimal classifier between the joint distribution $p(\normalfont{\textbf{H}, \textbf{s}})$ and $p(\normalfont{\textbf{H}})p(\normalfont{\textbf{s}})$ is capped at $Er^* = \frac{1}{2}\sum^{\normalfont{|\textbf{H}|}}_{g=1}p(\normalfont{\textbf{s}^g})^2$.
\end{lemma}
Based on our analysis in Section \ref{rethink_gcl}, we assume $\textbf{s}$ is a constant summary vector $\epsilon \textit{\textbf{I}}$, where $\epsilon$ is the constant in $\textbf{s}$. In addition, we assume $\epsilon$ in $\textbf{s}$ is independent from $p(\textbf{H})$. Then, we can derive the following lemma:
\begin{lemma}
\label{le:lem 1/2}
We assume $\normalfont{\textbf{s}}$ is a constant summary vector $\epsilon \textbf{\textit{I}}$ and $\epsilon$ of $\normalfont{\textbf{s}}$ is independent from $p(\normalfont{\textbf{H}})$, the error rate for the optimal classifier between the joint distribution $p(\normalfont{\textbf{H}, \textbf{s}})$ and the product of marginals $p(\normalfont{\textbf{H}})p(\normalfont{\textbf{s}})$ is $Er^* = \frac{1}{2}$.
\end{lemma}}

\change{\textit{Proof.} As $\epsilon$ is independent from $p(\textbf{H})$, we can see $p(\textbf{s})$ is independent from $p(\textbf{H})$. Thus, the joint distribution $p(\textbf{H}, \textbf{s})$ equals to the product of marginals $p(\textbf{H})p(\textbf{s})$. As a result, every sample from the joint is also a sample from the product of marginals. In this case, no classifier performs better than random guess, i.e., no classifier can discriminate samples from $p(\textbf{H}, \textbf{s})$ and $p(\textbf{H})p(\textbf{s})$ in this case. Therefore, we prove that $Er^* = \frac{1}{2}$ \cite{velivckovic2018deep}.}

\change{Then we present Theorem \ref{th: mi}, which is presented in DGI\cite{velivckovic2018deep}:
\begin{theorem}
\label{th: mi}
Define $\normalfont{\textbf{s}^*}$ as the optimal summary vector under the classification error of an optimal classifier between  $p(\normalfont{\textbf{H}, \textbf{s}})$ and $p(\normalfont{\textbf{H}})p(\normalfont{\textbf{s}})$. $\normalfont{\textbf{s}^*} = argmax_{\normalfont{\textbf{s}}}MI(\normalfont{\textbf{H};\textbf{s}})$, where MI stands for mutual information. 
\end{theorem}
Based on Theorem \ref{th: mi}, in DGI, they claim that for finite input sets and appropriate deterministic functions, minimising the classification error in the discriminator $\mathcal{D}(\cdot)$ (as shown in Equation \ref{eq:discriminator}) can be used to maximise the MI between the input and output of $\mathcal{R}(\cdot)$. However, under the aforementioned assumptions, the error rate $Er^*$ is a constant, and it is not practical to minimise the classification error. In addition, as $p(\textbf{s})$ is independent from $p(\textbf{H})$, we know $MI(\textbf{H};\textbf{s}) = 0$. Thus, we can see these findings contradict to Theorem \ref{th: mi}.}

\change{Instead of maximising the $MI(\textbf{H};\textbf{s})$, in this case, the discriminator $\mathcal{D}(\cdot)$ is responsible for maximising the similarity between positive embeddings and the constant summary vector $\textbf{s}$, while minimising the similarity between negative embeddings and $\textbf{s}$. This operation is equivalent to maximising the Jensen-Shannon divergence between the distribution of positive embeddings and negative embeddings. We show a theorem to explain this as follows:
\begin{theorem}
\label{th: optimise}
Assuming $\normalfont{\textbf{s}}$ is a constant summary vector $\epsilon \textit{\textbf{I}}$ and $\epsilon$ of $\normalfont{\textbf{s}}$ is independent from $p(\normalfont{\textbf{H}})$. Given a graph $\mathcal{G}$, a corrupted graph $\tilde{\mathcal{G}}$, and a GNN encoder $g_{\theta}(\cdot)$, we consider the distribution of positive embeddings $g_{\theta}(\mathcal{G})$ as $P^\mathbf{h}_{pos}$ and negative embeddings $g_{\theta}(\tilde{\mathcal{G}})$ as $P^\mathbf{h}_{neg}$. Optimising the DGI loss is equivalent to maximising the Jensen-Shannon divergence between $P^\mathbf{\hat{h}}_{pos}$ and $P^\mathbf{\hat{h}}_{neg}$, where $\mathbf{\hat{h}}$ is linearly transformed $\mathbf{h}$.
\end{theorem} 
\textit{Proof.} We first present the DGI loss with a constant summary vector:
\begin{equation}
\begin{aligned}
           \mathcal{L} &= \mathbb{E}_{\textbf{h} \sim P^\textbf{h}_{pos}} log \mathcal{D}(\textbf{h}, \textbf{s}) + \mathbb{E}_{\textbf{h} \sim P^\textbf{h}_{neg}} log(1 - \mathcal{D}(\textbf{h}, \textbf{s})), \\
           &=  \mathbb{E}_{\textbf{h} \sim P^\textbf{h}_{pos}} log (\textbf{h} \cdot \textbf{W} \cdot \textbf{s}) + \mathbb{E}_{\textbf{h} \sim P^\textbf{h}_{neg}} log(1 - \textbf{h} \cdot \textbf{W} \cdot \textbf{s}), \\
           &=  \mathbb{E}_{\textbf{h} \sim P^\textbf{h}_{pos}} log (\textbf{h} \cdot \textbf{W} \cdot \epsilon) + \mathbb{E}_{\textbf{h} \sim P^\textbf{h}_{neg}} log(1 - \textbf{h} \cdot \textbf{W} \cdot \epsilon),
\end{aligned}
\end{equation}
where $\textbf{h}$ is node embedding and $\textbf{W}$ is learnable weight matrix. Here, we consider $\textbf{h} \cdot \textbf{W}$ (i.e., linearly transformed $\textbf{h}$) as $\hat{\textbf{h}}$. Then, we consider the distribution of $\hat{\textbf{h}}$ generated with positive samples $\textbf{h}$ as $P^{\hat{\textbf{h}}_{pos}}$ and the distribution of $\hat{\textbf{h}}$ with negative samples $\textbf{h}$ as $P^{\hat{\textbf{h}}_{pos}}$.
Then, the equation becomes:
\begin{equation}
    \begin{aligned}
               \mathcal{L} &=  \mathbb{E}_{\hat{\textbf{h}} \sim P^{\hat{\textbf{h}}}_{pos}} log (sum(\epsilon \hat{\textbf{h}})) + \mathbb{E}_{{\hat{\textbf{h}}} \sim P^{\hat{\textbf{h}}}_{neg}} log(1 - sum(\epsilon \hat{\textbf{h}})), \\
            &=  \mathbb{E}_{\hat{\textbf{h}} \sim P^{\hat{\textbf{h}}}_{pos}} log (\epsilon \cdot agg(\hat{\textbf{h}})) + \mathbb{E}_{\hat{\textbf{h}} \sim P^{\hat{\textbf{h}}}_{neg}} log(1 - \epsilon \cdot agg(\hat{\textbf{h}})),
    \end{aligned}
\end{equation}
the above equation is very similar to Equation \ref{eq:loss js} and the only difference is there is a $\epsilon$ multiply with the $agg(\cdot)$ output. Here, $agg(\cdot)$ is summation. Thus, the proof for Theorem \ref{th: mi} still holds and prove Theorem \ref{th: optimise}.}

\change{Based on Theorem \ref{th: optimise}, we can see group discrimination without the summary vector is doing the same thing as DGI with a constant summary vector (i.e., separating positive and negative distribution). Thus, we are motivated to remove the summary vector from the loss and proposed the group discrimination loss in Equation \ref{eq:binary DGI}.}

\change{
Instead of relying on a summary vector $\textbf{s}$ to discriminate positive and negative samples in $\textbf{H}$ (i.e., calculating their similarity with the summary vector), we directly use a binary cross entropy loss to classify these samples. 
Removing the summary vector $\textbf{s}$ is beneficial to the computation efficiency because it eases the burden of gradient computation, e.g., to compute the gradient for $\textbf{s}$, we need to store and use all the parameters in the model to conduct backward propagation. However, in group discrimination, we do not need the summary vector and only aggregate node embeddings to obtain prediction.}


%% file: algorithm.tex

\IncMargin{1em}
\begin{algorithm}
\SetKwData{Left}{left}\SetKwData{This}{this}\SetKwData{Up}{up}
\SetKwFunction{Union}{Union}\SetKwFunction{FindCompress}{FindCompress}
\SetKwInOut{Input}{Input}\SetKwInOut{Output}{Output}
\Input{Input Graph  $\mathcal{G} = \{\textbf{X} \in \mathbb{R}^{N\times D}, \textbf{A} \in \mathbb{R}^{N\times N}\}$; GNN encoder $g_{\theta}(\cdot)$; Projector $f_{\theta}(\cdot)$; Number of nodes $N$; Number of feature dimensions $D$; Number of hidden dimensions $D'$; Number of training epochs $T$}
\Output{Final representation $\textbf{H}$}
\BlankLine
\emph{//Model Training}\;
\For{$t = 1$ to $T$}{
\emph{//Augmentation}(optional)\;
Conduct feature dropout and edge dropout on $\mathcal{G}$ to obtain $\hat{\mathcal{G}} = \{\hat{\textbf{X}}, \hat{\textbf{A}}\}$\;
\emph{//Corruption}\;
Corrupt $\hat{\mathcal{G}}$ to obtain $\tilde{{\mathcal{G}}} = \{\tilde{\textbf{X}}, \tilde{\textbf{A}}\}$\;
\emph{//Compute Encoding}\;
Input $\hat{\mathcal{G}}$ and $\tilde{{\mathcal{G}}}$ to  $g_{\theta}(\cdot)$ and $f_{\theta}(\cdot)$ to obtain graph embedding $\rm{\hat{\textbf{H}}_{\theta}}$ $= f_{\theta}(g_{\theta}(\hat{\mathcal{G}}))$ and $\rm{\tilde{\textbf{H}}_{\theta}}$ $= f_{\theta}(g_{\theta}(\tilde{\mathcal{G}}))$, respectively\;
\emph{//Aggregation}\;
Concatenate $\rm{\hat{\textbf{H}}_{\theta}}$ and $\rm{\tilde{\textbf{H}}_{\theta}}$ to obtain $\rm \bar{\textbf{H}}_{\theta}$\;
Conduct aggregation on $\rm \bar{\textbf{H}}_{\theta} \in \mathbb{R}^{2N \times D'}$ to obtain the prediction vector $\hat{\textbf{y}}_{\theta} \in \mathbb{R}^{2N}$\;

\emph{//Compute Loss}\;
Calculate loss $\mathcal{L} = -\frac{1}{2N}(\sum_{i = 1}^{2N}y_i\log \hat{y}_i + (1 - y_i)\log(1 - \hat{y}_i))$, where $\hat{y}_i \in \mathbb{R}^{1}$ is the prediction for one node sample and $\hat{y}_i \in \hat{\textbf{y}}_{\theta}$\;
\emph{//Update parameters}\;
Update trainable parameters in $g_{\theta}(\cdot)$\ and $f_{\theta}(\cdot)$\;
}
\emph{//Inference}\;
Obtain local embedding $\rm{\textbf{H}_{\theta}}$ $= f_{\theta}(g_{\theta}({\mathcal{G}}))$ \;
Obtain global embedding $\textbf{H}_{\theta}^{global} = \textbf{A}^n \textbf{H}_{\theta}$\;
Obtain final embeddings for downstream tasks $\textbf{H} =  \textbf{H}_{\theta}^{global} + \textbf{H}_{\theta}$\;
\caption{The Overall Procedure of GGD}
\label{alg}
\end{algorithm}\DecMargin{1em}